\documentclass[a4paper, amsfonts, amssymb, amsmath, reprint, showkeys, nofootinbib, twoside, onecolumn, notitlepage]{revtex4-1}
\usepackage[english]{babel}
\usepackage[utf8]{inputenc}
\usepackage[colorinlistoftodos, color=green!40, prependcaption]{todonotes}
\usepackage{amsmath}
\makeatletter\AtBeginDocument{\let\@elt\relax}\makeatother
\usepackage{natbib}
\usepackage{algorithm} 
\usepackage{algorithmic} 
\usepackage{comment}
\usepackage[toc,page]{appendix}
\usepackage{nccmath}

\usepackage{verbatim} 

\begin{document}

\title[Models of Knowledge Distillation]{Solvable Model for Inheriting the Regularization \\ through  Knowledge Distillation}




\author{Luca Saglietti}
    \email{luca.saglietti@gmail.com} 
    
\author{Lenka Zdeborov\'a}
    \affiliation{SPOC laboratory, EPFL, Switzerland.}


\begin{abstract}%
In recent years the empirical success of transfer learning with neural networks has stimulated an increasing interest in obtaining a theoretical understanding of its core properties. Knowledge distillation where a smaller neural network is trained using the outputs of a larger neural network is a particularly interesting case of transfer learning. In the present work, we introduce a statistical physics framework that allows an analytic characterization of the properties of knowledge distillation (KD) in shallow neural networks. Focusing the analysis on a solvable model that exhibits a non-trivial generalization gap, we investigate the effectiveness of KD. We are able to show that, through KD, the regularization properties of the larger teacher model can be inherited by the smaller student and that the yielded generalization performance is closely linked to and limited by the optimality of the teacher. Finally, we analyze the double descent phenomenology that can arise in the considered KD setting.  %
\end{abstract}

%


\maketitle

\section{Introduction}
\label{sec:KD}

Deep learning practice in the past decade has repeatedly confirmed a remarkable observation: Stochastic Gradient Descent (SGD) based training of neural networks becomes easier and more effective as the number of tunable parameters increases. While a higher model complexity could in principle entail high risks of over-fitting, large scale Deep Neural Networks (DNNs) display surprising generalization capabilities, allegedly allowed by an “implicit regularization” mechanism \cite{neyshabur2014search,zhang2016understanding} that still escapes clear theoretical understanding. On the flip side, the steady scaling up of DNN architectures also carries a massive increase in associated inference and memory costs. 

Many attempts at achieving better quality-computation trade-offs have been proposed in the last decade \cite{han2015learning, han2015deep, jacob2018quantization, frankle2018lottery}, often based on the observation that good generalization scores can be retained if one first trains a more complex DNN and then derives a lighter model from it. In this context, Knowledge Distillation (KD) \cite{hinton2015distilling} has established itself as one of the most popular transfer learning and network compression strategies \cite{anil2018large, chen2018darkrank, chen2017learning, yim2017gift, yu2017visual, kim2016sequence}. The general idea of KD is to try and transfer the generalization properties from a larger capacity model (teacher) to a weaker model (student) at training: instead of learning directly from the vector encodings of the ground truth labels, the KD student learns from the outputs (``dark knowledge'') produced by the teacher model on the same training dataset. Not only the KD optimization step, with real-valued outputs, seems to be generally more well-behaved than usual training, numerical experiments also show that with very little fine-tuning at the level of the employed regularization and optimization heuristics one can reach competitive generalization scores \cite{tang2020understanding}.

Despite its effectiveness, KD is still not very well understood from a theoretical standpoint. Few recent works \cite{celik2017patient, phuong2019towards, tang2020understanding, rahbar2020unreasonable, yuan2019revisit, furlanello2018born} have attempted to analyse KD in controlled settings, breaking down its net positive impact into separate contributions and proposing a connection between the effectiveness of KD and the role of label smoothing and sample reweighting strategies and that of priors on data geometry \cite{yuan2019revisit, furlanello2018born}. In this work we approach the problem from a statistical physics perspective, aiming to characterize the typical generalization performance achieved by a KD student in the asymptotic limit of large input dimension and dataset size. The main questions we want to investigate are how KD can transfer the regularization properties between mismatched models and when the KD student can display an improvement upon the best score achievable with usual training procedures. 

In order to allow a mathematical definition of knowledge distillation, consider a typical classification problem where, given a large dataset of input-output associations $\mathcal{D}=\{\boldsymbol{x}^\mu,y^\mu\}_{\mu=1}^M$ the task is to learn a parametrized rule $f(\boldsymbol{x}^\mu, \left\{\boldsymbol{w}\right\})$ ($f$ representing a neural network and $\{\boldsymbol{w}\}$ its parameters or weights) that allows correct classification of test data points not seen in the training set.
As customary, learning can be framed as an empirical risk minimization problem, introducing a regularized loss-function:
\begin{equation} 
\mathcal{L}(\left\{\boldsymbol{w}\right\}, \mathcal{D})=\sum_{\mu=1}^{M}\ell\left(y^{\mu}, \sigma\left(f(\boldsymbol{x}^\mu, \left\{\boldsymbol{w}\right\})\right)\right)+\frac{\lambda}{2}\lVert \boldsymbol{w} \rVert \label{eq:loss}
\end{equation}
where, in the binary classification case, $\sigma(\cdot)$ is the sigmoid activation function $\sigma\left(x\right)=\left(1+\exp\left(-x\right)\right)^{-1}$ ($softmax$ in the multi-class case), and where a typical choice would be a cross-entropy loss $\ell(p,q) = \mathcal{H}(p,q)$:
\begin{equation}
\mathcal{H}(p,q) = -p \log q -(1-p) \log(1-q)
\end{equation}
and an $L_2$-norm regularization.

In knowledge distillation (KD), one assumes to be granted access to the outputs $\tilde{f}(\boldsymbol{x}^\mu, \left\{\tilde{\boldsymbol{w}}\right\})$ produced over the training set by a (more complex) teacher model $\{ \tilde{\boldsymbol{w}}\}$, and one aims at training a (weaker) student model through the modified loss:
\begin{equation}
\mathcal{L}^\prime(\left\{\boldsymbol{w}\right\}; \left\{ \boldsymbol{\tilde w}\right\},\mathcal{D})=\sum_{\mu=1}^{N}\ell^\prime \left(y^{\mu}, \tilde{f}(\boldsymbol{x}^\mu, \left\{\tilde{\boldsymbol{w}}\right\}), f(\boldsymbol{x}^\mu, \left\{\boldsymbol{w}\right\}),\chi\right)+\frac{\lambda}{2}\lVert \boldsymbol{w} \rVert
\end{equation}
where:
\begin{equation}
\ell^\prime \left(y, p, q, \chi \right)= (1-\chi) \mathcal{H}(y,\sigma(q)) + \chi\, \mathcal{H}(\sigma(p),\sigma(q)). \label{eq:distillation_loss}
\end{equation}

The student is thus mixing the usual data-fitting approach with a goal of approximating the behavior of the teacher, the external parameter $\chi$ serving to balance between fitting the ground truth labels and the teacher outputs. A key idea behind distillation is that for the student the optimization process becomes more transparent, as it can rely on the more explicit knowledge derived from the teacher outputs: the softer outputs can prevent student overconfidence on noisy data points and highlight correlations among different labels. Note that, in multi-class problems one typically considers also a distillation temperature $T$, reweighting teacher and student outputs: we will ignore this additional external parameter in the following, since we focus on a binary classification setting (a short analysis of its effect can be found in Appendix\,\ref{sec:more_on_inheritance}).

\subsection{Main contributions}

In this manuscript, we develop and apply an analytic framework to study knowledge distillation in models where the learning performance is solvable with the replica method. We then consider specifically a Gaussian mixture model where the student is constrained to be sparse, and we perform a series of controlled studies that allowed an investigation of the inheritance properties of knowledge distillation. All analytical results are crosschecked with numerical experiments. Our main results can be summarized in the following qualitative observations:
\begin{itemize}
    \item Without any fine-tuning at the level of the student loss function, using KD allows a transfer of the (possibly fine-tuned) regularization properties of the teacher, even if the two models are mismatched and even if the regularization strategy in the teacher training is not known explicitly.
    \item When the regularization mechanism employed for regularizing the teacher can also be applied directly to the student, fine-tuning the direct regularization and fine-tuning the parameters of the KD loss leads to comparable generalization performance. No improvement is observed in this setting.
    \item If one can access a trained network with superior generalization performance and employ it as a teacher in a KD process, also the KD student will inherit superior generalization properties.  
    \item In the limit of zero direct regularization on the student, the KD loss gives rise to a hybrid double-descent phenomenology, displaying both logistic regression and linear regression types of cusps.
\end{itemize}

Of course, the results we derived in the simple Gaussian mixture model may not generalize directly to more complex network architectures or different types of model mismatches. We argue, however, that the observed qualitative behavior is in line with the empirical observations about KD practice in deep learning \cite{hinton2015distilling,rahbar2020unreasonable}, and support the propositions that transferring knowledge from larger (implicitly regularized) neural network models is almost automatically beneficial for the test performance of weaker students. Moreover, the development of a general theoretical framework for this type of study could stimulate a similar analysis in more realistic settings. 

In the next section, we introduce our analytical framework, yielding an asymptotic description of training through knowledge distillation. In section \ref{sec:GMM}, we present a solvable model where the test performance associated with logistic regression is largely sub-optimal and we define in which sense the considered student network is a smaller model than the teacher. In section \ref{sec:applied}, we apply the analytical framework to the model and we derive a set of deterministic fixed-point equations that allow an estimate of the KD student generalization performance. In section \ref{sec:main_results}, we showcase the main results of this work, comparing our predictions with numerical simulations. In particular, we characterize the inheritance properties and the limits of KD and show when KD can potentially lead to improved generalization with respect to typical logistic regression. In section \ref{sec:double_descent}, we focus on the double-descent phenomenology that appears in our simple transfer learning setup. Finally, in section \ref{sec:conclusions} we discuss our results and propose some future research perspectives.

\section{Statistical physics framework to analyze knowledge distillation}
\label{sec:StatPhysFramework}

The main technical contribution of this work consists in the introduction of an analytic framework based on the replica formalism \cite{mezard1987spin, mezard2009information}, that allows the characterization of learning through knowledge distillation in tractable models. 

The proposed analytic setup stems from the simple observation that KD can naturally be framed as a 2-level problem: in the first step one trains a teacher model with the true labels and the dataset $\mathcal{D}$; in the second step a student network is trained with the same inputs and the outputs produced by the teacher. From the perspective of the replica method, the fact that the two systems are sharing the same inputs (same quenched disorder) is effectively coupling them. However, because of the fixed order of the two training procedures, the teacher model is not affected by the presence of the student and therefore its statistical properties can be determined self-consistently. On the other hand, the statistical measure of the student is directly dependent on the specific realization of the teacher.  

In particular, we can characterize the typical KD student by considering the disordered partition function:
\begin{equation}
Z(\{ \tilde{\boldsymbol{w}} \}, \mathcal{D})=\lim_{\beta\to\infty}\int d \boldsymbol{w} \,\,e^{-\beta \mathcal{L}^\prime(\{\boldsymbol{w}\};\{\boldsymbol{\tilde{w}}\}, \mathcal{D})}
\end{equation}
in the limit $\beta\to\infty$, where the measure focuses on the minimizers of the KD loss functions, and then evaluate the free-entropy $\Phi$ of the model by performing an external average over the realization of the dataset and an internal average over the measure of the trained teacher: 
\begin{equation}
\Phi = \frac{1}{N}\left<\,\left< \,\,\log Z(\{ \tilde{\boldsymbol{w}} \}, \mathcal{D})\,\,\right>_{\{ \tilde{\boldsymbol{w}} \}} \right>_{\mathcal{D}}. \label{eq:double_average}
\end{equation}
Quantities of interest such as the value of test error of the student are then readily derived from this free entropy in ways standard to statistical physics \cite{mezard2009information}. Obtaining a close formula for the free entropy $\Phi$ is hence the key difficulty that can be overcome using the replica trick \cite{mezard1987spin}. The replica formalism required for evaluating the double average in Eq.\,(\ref{eq:double_average}) is equivalent to a Franz-Parisi potential computation \cite{franz1997phase}, where one first samples a configuration from an independent equilibrium measure and than evaluates the free-energy of a coupled system sharing the same realization of the disorder. 

The computation starts from a chain of identities based on two separate replica tricks:
\begin{eqnarray}
\Phi=\frac{1}{N}\left< \,\left< \,\,\log \lim_{\beta\to\infty}\int d \boldsymbol{w} \,\,e^{-\beta \mathcal{L}^\prime(\{\boldsymbol{w}\};\{\tilde{\boldsymbol{w}}\}, \mathcal{D})}\,\,\right>_{\{ \tilde{\boldsymbol{w}} \}} \right>_{\mathcal{D}} = \\ 
\frac{1}{N}\left< \,\left< \,\,\lim_{n\to 0} \frac{\partial}{\partial n}\,\, \lim_{\beta\to\infty}\int \prod_{a=1}^n d \boldsymbol{w}^a \,\,e^{-\beta \mathcal{L}^\prime(\{\boldsymbol{w}^a\};\{\tilde{\boldsymbol{w}}\}, \mathcal{D})}\,\,\right>_{\{ \tilde{\boldsymbol{w}} \}} \right>_{\mathcal{D}} = \\
\frac{1}{N}\left< \, \lim_{\tilde{n},n\to0} \frac{\partial}{\partial n}\,\, \lim_{\tilde{\beta},\beta\to\infty} \int \prod_{c=1}^{\tilde{n}} d \tilde{\boldsymbol{w}}^c e^{-\tilde{\beta} \mathcal{L}(\{\tilde{\boldsymbol{w}}^c\}; \mathcal{D})} \int \prod_{a=1}^n d \boldsymbol{w}^a \,\,e^{-\beta \mathcal{L}^\prime(\{\boldsymbol{w}^a\};\{\tilde{\boldsymbol{w}}^1\}, \mathcal{D})}\, \right>_{\mathcal{D}}. \label{eq:fp-formula}
\end{eqnarray}
In order to evaluate the disorder average, in the second line the logarithm is removed by replicating the student configuration $\{ \boldsymbol{w}^a\}_{a=1}^n$ (using the identity $\log x = \lim_{n\to 0} \partial_n x^n $). In the third line, instead, the average over the teacher is removed by introducing $\tilde{n}-1$ non-interacting and a single interacting replica of the teacher $\{ \tilde{\boldsymbol{w}}^c\}_{c=1}^{\tilde{n}}$, so that in the $\tilde{n}\to0$ limit one can recover the expectation over its measure.     

Because of concentration properties in high-dimensions, the coupled free-entropy asymptotically converges to a deterministic function of a narrow set of order parameters that capture the geometrical distribution of teacher and student configurations. Enforcing a saddle-point condition for the free-energy allows the derivation of a system of fixed-point equations that can yield an asymptotic prediction for these order parameters, to be compared with the results of numerical simulations. 

The proposed formalism is general and may be applied to analyze knowledge distillation in any learning model which is amenable of a description through the replica method. Note that the entailed computation is quite standard in statistical physics and is believed to be exact, although non-rigorous in general. Moreover, an important remark is that there currently are strong technical limitations which restrict the set of models tractable with the replica method to a class of shallow network architectures \cite{barbier2019optimal,aubin2018committee}, but these limitations might be lifted with future progress in the field.   

\section{Gaussian Mixture Model}
\label{sec:GMM}

We now provide a brief introduction to a model recently analyzed in \cite{mignacco2020role} with the replica method, which will be employed as a prototypical study case in the rest of the paper. The same models can be studied with other tools, some of them rigorous, but here we focus on the replica solution of the model because that is the one that can readily be extended to analyze the knowledge distillation along the lines of section~\ref{sec:StatPhysFramework}. We consider a high-dimensional binary classification problem where data is generated according to a Gaussian mixture and the learning model is a linear classifier trained through $L_{2}$-regularized logistic regression. 

In particular, let $N$ denote the input dimension and $M$ denote the size of the training set $\mathcal{D}$. We assume the data points in $\mathcal{D}$ to be Gaussian distributed around two centroids, located on the same axis and positioned respectively at $\pm \frac{\boldsymbol{v}}{\sqrt{N}}$, $\boldsymbol{v} \in \mathbb{R}^N$. Moreover, we assume the two clusters to contain respectively a fraction $\rho$ and $(1-\rho)$ of the points. Thus, data $\{\boldsymbol{x}^\mu, y^\mu \}_{\mu=1}^M$ is generated according to the process:
\begin{equation}
\boldsymbol{x}^{\mu}=\left(2y^{\mu}-1\right)\frac{\boldsymbol{v}}{\sqrt{N}}+\sqrt{\Delta}\boldsymbol{z}    
\end{equation}
where each component of the signal $\boldsymbol{v}$ and noise $\boldsymbol{z}$ is i.i.d. Gaussian, $v_i,z_i\sim \mathcal{N}(0,1)$. The binary labels $y_\mu\in\{0,1\}$ that determine the cluster membership of the points follow the skewed distribution $y^{\mu}\sim\rho\,\delta\left(y^{\mu}-1\right)+(1-\rho)\,\delta\left(y^{\mu}\right)$. We also specialize to the case of a single layer network, with: 
\begin{equation}
    f(\boldsymbol{x}^\mu, \left\{\boldsymbol{w}\right\}) = \frac{\boldsymbol{x}^\mu \cdot \boldsymbol{w}}{\sqrt{N}} + b
\end{equation}
where the weights $\boldsymbol{w}\in\mathbb{R}^N$ and the bias $b\in\mathbb{R}$ represent the tunable parameters of the model. Note that, as soon as the training set is no longer linearly separable, the optimal learning strategy is to try and align the weights in the direction of the signal $\boldsymbol{v}$, so that the probability of a correct labeling of the data points is maximized.   

Non-trivial behaviour was described in this model in the scaling limit where both $N,M\to\infty$, while their ratio $\alpha=M/N$ remains of $\mathcal{O}(1)$ \cite{mignacco2020role}. Asymptotically, the model is fully solvable and one can characterize the typical learning performance as a function of the model parameters and of the regularization intensity $\lambda$. For the reader's convenience we report some results in Appendix\,\ref{sec:replica_computations}, but we reference \cite{mignacco2020role} for the full details on the properties of this model. 

\subsection{Sub-optimal performance of logistic regression}

One of the main motivations for considering this simple model in the present knowledge distillation study comes from the sub-optimal generalization behavior of regularized logistic regression in the unbalanced cluster case, at $\rho<0.5$. As reported in \cite{mignacco2020role}, given the overlap between the weight configuration and the signal, $m=\frac{\boldsymbol{w}\cdot\boldsymbol{v}}{N}$ and the norm $q=\frac{\boldsymbol{w}\cdot\boldsymbol{w}}{N}$, one obtains the asymptotic generalization error of the trained configuration from the following analytic expression:
\begin{equation}
\epsilon_g = \rho H\left(\frac{m+b}{\sqrt{\Delta q}}\right) +  (1-\rho) H\left(\frac{m-b}{\sqrt{\Delta q}}\right). \label{eq:gen_error}
\end{equation}
This score can then be compared with the Bayes optimal generalization error (computed by matching the inference model with the generative one), which represents a lower bound for the performance of any learning algorithm. 
\begin{figure}[ht]
\centering
\includegraphics[width=0.5\textwidth]{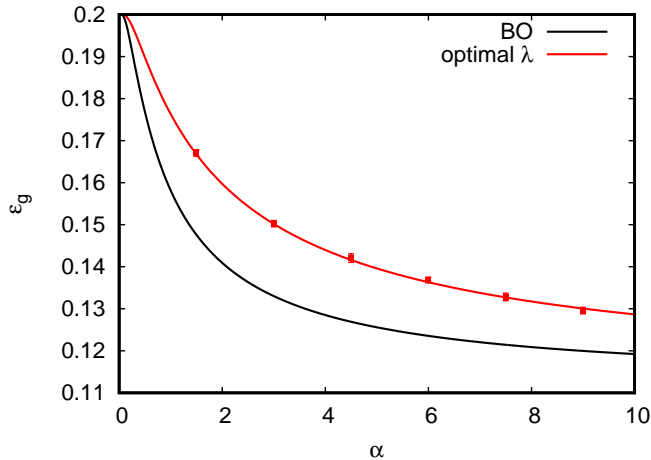}
\caption{\label{fig:gen_GAP} Generalization performance of $L_2$-regularized logistic regression compared to the Bayes optimal lower bound, in a Gaussian Mixture with $\rho=0.2$ and $\Delta=1$. Red line: regularized logistic regression with optimal intensity $\lambda$. Black line: Bayes optimal performance. The data points with error bars represent the results of numerical experiments at $N=4000$ ($10$ samples per point).}
\end{figure}

Remarkably, in this specific model there always exist at least a point estimator that achieves such Bayes-optimal performance, constructed according to a Hebbian principle \cite{hebb2005organization}:
\begin{equation}
     \boldsymbol{w}_{BO} = \frac{1}{\alpha \sqrt{N}}\sum_{\mu=1}^{\alpha N} (2y_\mu-1) \boldsymbol{x}_\mu , \,\,\,\,\,\,\,\, b_{BO}=\frac{\Delta \lVert\boldsymbol{w}_{BO} \rVert^2_2}{2}\log\frac{\rho}{1-\rho}.    \label{eq:plug-in}
\end{equation}
Thus, the question is whether one can set an optimal value of the regularization such that logistic regression can perform similarly. The somewhat surprising answer is that, at $\rho<0.5$ and any value of $\alpha$ and $\Delta$, one observes a sizable gap between the best generalization performance obtained through logistic regression and the optimal one. This phenomenon can be clearly seen in Fig.\,\ref{fig:gen_GAP}. In section \ref{sec:main_results} we will investigate whether KD can help the student close this performance gap.
Note that this sub-optimal behavior does not appear with balanced clusters $\rho=0.5$, where the optimal regularization level is obtained in the limit $\lambda\to\infty$ \cite{mignacco2020role}.

\subsection{Teacher-student mismatch}

In the present work, we want to analyze a setting where teacher and student model classes are mismatched, so that the weaker student is not able to exactly replicate the behavior of the teacher. Introducing some type of mismatch is not only closer to KD practice, but also crucial for inducing a richer model phenomenology. As a matter of fact, it was shown in \cite{phuong2019towards} that as soon as the number of constraints becomes larger than $\alpha>1$ a linear KD student can trivially recover the teacher weight configuration. 

We thus consider a scenario where the student model can only train a fraction $0<\eta<1$ of its weights while the rest is set to $0$ \emph{ab initio}. In this way it will be impossible for the student to exactly infer (and achieve the same performance as) the teacher, and we can focus on the transfer of knowledge between the two models.

Note that, because of the simple nature of the considered generative model, setting a fraction $\eta<1$ is equivalent to rescaling the effective signal-to-noise ratio in the student learning problem. In fact, the two inference tasks with $\left\{\eta,\alpha,\Delta\right\}$ and with $\left\{1,\alpha/\eta,\Delta/\eta\right\}$ are information-theoretically equivalent in the asymptotic limit. Moreover, one can easily define a Bayes optimal lower bound also in the sparse sub-space spanned by the student, achieved by a point-estimator with the same bias $b_{\rm BO}$ as in Eq.\,(\ref{eq:plug-in}) and trimmed weights:
\begin{equation}
\left(w_{\rm BO}\right)_{i}=\left\{\begin{array}{lr}
        \frac{1}{\alpha\sqrt{N}} \sum\left(2y^{\mu}-1\right)x_{i}^{\mu}, & \text{for } i\le\eta N\\
        0, & \text{for } i>\eta N
        \end{array}\right. . \label{eq:sparse_plugin}
\end{equation}
As expected, one can easily prove that the associated performance coincides with the typical Bayesian generalization obtained after rescaling $\alpha$ and $\Delta$ by a factor $1/\eta$.

\section{Knowledge Distillation in the Gaussian Mixture Model}
\label{sec:applied}

We apply the replica framework sketched in Sec.~\ref{sec:StatPhysFramework} to derive a set of deterministic equations characterizing typical knowledge distillation processes in the above introduced logistic regression setting, where one first trains a teacher linear classifier and then employs the KD loss Eq.\,(\ref{eq:distillation_loss}) to train a sparsified linear student. The convex nature of the two nested optimization problems justifies the employment of the so-called Replica Symmetric ansatz \cite{mignacco2020role}, which simplifies the analysis considerably. 
As the replica computation is still quite involved, we defer a detailed description to the Appendix, and report here the obtained final expressions. 

We remind that the main parameters of the setting we analyze are the noise variance $\Delta$ and the label fraction $\rho$, the number of samples per dimension $\alpha=M/N$ and the student-sparsity level $\eta$. Specializing Eq.\,(\ref{eq:fp-formula}) to our study case we get:
\begin{equation}
\Phi=\frac{1}{N}\lim_{n,\tilde{n}\to0}\partial_{n}\left<\lim_{\tilde{\beta}\to\infty}\lim_{\beta\to\infty}\int\prod_{c=1}^{\tilde{n}}d \tilde{\boldsymbol{w}}^{c} e^{-\frac{\beta\tilde{\lambda}}{2}\lVert \tilde{\boldsymbol{w}}^c \rVert^2} \int\prod_{c=1}^{\tilde{n}}d\tilde{b}^{c}\prod_{\mu,c}e^{-\frac{\tilde{\beta}}{2}\ell\left(y^{\mu},\sigma\left(\sum_{i=1}^{N}\frac{\tilde{w}_{i}^{c}x_{i}^{\mu}}{\sqrt{N}}+\tilde{b}^{c}\right)\right)}\right. \nonumber
\end{equation}
\begin{equation}
\left.\times\int\prod_{a=1}^{n}d \boldsymbol{w}^{a} e^{-\frac{\beta\lambda}{2}\lVert \boldsymbol{w}^a \rVert^2} \int\prod_{a=1}^{n}db^{a}\prod_{\mu,a}e^{-\frac{\beta}{2}\ell^\prime \left(y^\mu,\sigma\left(\sum_{i=1}^{N}\frac{\tilde{w}_{i}^{1}x_{i}^{\mu}}{\sqrt{N}}+\tilde{b}^{1}\right),\sigma\left(\sum_{i=1}^{N}\frac{w_{i}^{a}x_{i}^{\mu}}{\sqrt{N}}+b^{a}\right),\chi\right)}\right>_{\left\{ \boldsymbol{x}^{\mu},y^{\mu}\right\} },
\end{equation}
where $b, \tilde{b}$ are the biases, and $\lambda, \tilde \lambda$ the strength of the regularization of the student and teacher. 

In order to perform disorder average and remove the dependency on the specific realization of the Gaussian mixture dataset, one can first isolate teacher and student preactivations by introducing the associated Dirac's $\delta$-functions. Then, it becomes possible to factorize over the samples and the input components and take the expectation over $x^\mu_i$. Finally, one can discard the $o(N^{-1})$ terms and obtain the effective interaction between the various replicas, mediated by a set of overlap order parameters.

Thus, in the replica symmetric assumption, the only relevant quantities that fully characterize the studied model are:
\begin{itemize}
\item The overlap between teacher $\boldsymbol{\tilde w}$ and student $\boldsymbol{w}$ weight configurations with the signal $\boldsymbol{v}$ denoted $\tilde{m}=\frac{\tilde{\boldsymbol{w}}\cdot\boldsymbol{v}}{N}$, $m=\frac{\boldsymbol{w}\cdot\boldsymbol{v}}{N}$.
\item The norms $\tilde{q}=\frac{\tilde{\boldsymbol{w}}\cdot\tilde{\boldsymbol{w}}}{N}$ and $q=\frac{\boldsymbol{w}\cdot\boldsymbol{w}}{N}$. 
\item The teacher-student overlap $S=\frac{\boldsymbol{w}\cdot \tilde{\boldsymbol{w}}}{N}$.
\item The vanishing variances $\delta \tilde{q}$, $\delta q$ and $\delta S$ (see Appendix\,\ref{sec:replica_computations} for a detailed definition), opportunely rescaled in the $\tilde\beta,\beta\to\infty$ limit.
\end{itemize}
and their associated conjugate variables, denoted with a hat symbol in the following. Then, after some calculations, one can express the free-entropy as an extremum operation:
\begin{equation}
\Phi=\underset{b, m, q, \delta q, S, \delta S, \hat{m}, \hat{q}, \delta \hat{q}, \hat{S}, \delta \hat{S}}{\mathrm{extr}}-\left(\hat{m}m+\frac{1}{2}\left(\hat{q}\delta q-\delta\hat{q}q\right)+\left(\hat{S}\delta S+\delta\hat{S}S\right)\right)+\eta\,g_{s}+\alpha\,g_{e}, \label{eq:dist_finalFE}
\end{equation}
where we introduced the entropic and energetic contributions:
\begin{equation}
g_{s}=\frac{1}{2}\frac{\left(\hat{m}+\hat{\tilde{m}}\frac{\delta\hat{S}}{\tilde{\lambda}+\delta\hat{\tilde{q}}}\right)^{2}+\hat{q}+2\frac{\delta\hat{S}\hat{S}}{\tilde{\lambda}+\delta\hat{\tilde{q}}}+\frac{\hat{\tilde{q}}\delta\hat{S}^{2}}{\left(\tilde{\lambda}+\delta\hat{\tilde{q}}\right)^{2}}}{\lambda+\delta\hat{q}}
\end{equation}
\begin{equation}
g_{e}=\left\langle \int\mathcal{D}z\int\mathcal{D}\tilde{z}\,M_{E}^{\star}\right\rangle _{y} \label{eq:stud_energetic_term}
\end{equation}
with $\mathcal{D}z$ and $\mathcal{D}\tilde{z}$ denoting independent normalized Gaussian measures and the average $\left< \cdot \right>_y$ taken over the distribution of the cluster labels. The argument of Eq.\,(\ref{eq:stud_energetic_term}) is obtained from a one-dimensional optimization problem:
\begin{equation}
M_{E}^{\star}=\max_{u}\left\{ -\frac{1}{2}u^{2}-\frac{1}{2}\ell^\prime\left(y,\sigma\left(\tilde{h}\left(\tilde{u}^{\star}\right)\right),\sigma\left(h\left(u,\tilde{u}^{\star}\right)\right),\chi\right)\right\} \label{eq:student_opt}
\end{equation}
where $\tilde{h}$ and $h$ represent the teacher's and student's output pre-activations, given respectively by:
\begin{equation}
\tilde{h}\left(\tilde{u}\right)=\sqrt{\Delta\delta\tilde{q}}\tilde{u}+\sqrt{\Delta\tilde{q}}\tilde{z}+\left(2y-1\right)\,\tilde{m}+\tilde{b}
\end{equation}
\begin{equation}
h\left(u,\tilde{u}\right)=\sqrt{\Delta\delta q}u+\sqrt{\Delta\left(q-\frac{S^{2}}{\tilde{q}}\right)}z+\sqrt{\Delta}\frac{\delta S}{\sqrt{\delta\tilde{q}}}\tilde{u}+\sqrt{\Delta}\frac{S}{\sqrt{\tilde{q}}}\tilde{z}+\left(2y-1\right)\,m+b
\end{equation}
\begin{equation}
\tilde{u}^{\star}=\mathrm{argmax}_{\tilde{u}} \left\{ -\frac{1}{2}\tilde{u}^{2}-\frac{1}{2}\ell\left[y,\sigma\left(\tilde{h}\left(\tilde{u}\right)\right)\right] \right\}. \label{eq:teacher_opt}
\end{equation}
Note that the 2-level structure of Knowledge Distillation clearly appears in the concatenated optimization entailed in Eqs.\,(\ref{eq:student_opt}) and (\ref{eq:teacher_opt}). 

Since the teacher measure does not depend on the student,
the value of the associated order parameters can be determined independently by optimizing a simpler free-entropy. The corresponding fixed-point equations are reported in Appendix\,\ref{sec:replica_computations} (and are equivalent to those presented in \cite{mignacco2020role}). 

Once the saddle-point values for the order parameters of the models are evaluated for a given set of parameters $\{\alpha, \Delta, \rho, \eta, \tilde{\lambda}, \lambda, \chi\}$, the corresponding generalization can be obtained via Eq.\,(\ref{eq:gen_error}).

In the present work we do not seek a rigorous proof of the replica predictions (e.g., following a similar Gordon Minimax approach as in \cite{mignacco2020role}). We will, however, provide numerical confirmation of the consistency of the analysis in the next section.

\section{Main Results}
\label{sec:main_results}

In this section we will consider a series of learning settings encompassed in the analytic framework in order to investigate the effectiveness and properties of knowledge distillation.

The external parameters of the studied model are the dataset size to input dimension ratio $\alpha$, the cluster spread $\Delta$, the relative size of the two clusters $\rho$, and the sparsity level of the student $\eta$. In the following we will focus on a representative case, with normal Gaussian noise $\Delta=1$, unbalanced clusters $\rho=0.2$ and a half-sparse student $\eta=0.5$, and explore various ranges of $\alpha$ (some experiments in the balanced case are reported in Appendix\,\ref{sec:balanced}).
We will also adjust the $L_2$ regularization intensity in the teacher and student losses, $\tilde{\lambda}$ and $\lambda$, and the KD mixing parameter, $\chi$ (see Eq.\,(\ref{eq:distillation_loss})), in order to evaluate the variation in the student generalization performance.  

\subsection{Inheriting the regularization}

In the first experiment, we study whether learning from the outputs produced by the teacher instead of the true labels can indirectly regularize the student network and improve its generalization performance. 

\begin{figure}[ht]
\centering
\includegraphics[width=0.5\textwidth]{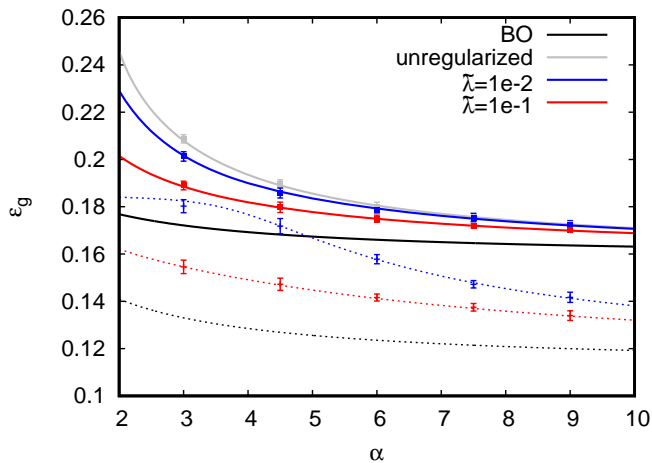}
\caption{ \label{fig:regularizations} Comparison between the generalization performance of a $\eta=0.5$ sparse student (full curves) and the corresponding ridge regularized  teacher (dashed curves), in a pure distillation setting ($\rho=0.2$, $\Delta=1$, $\chi=1$), as a function of $\alpha$. 
The data points with error bars represent the results of numerical experiments at $N=4000$ ($10$ samples per point). }
\end{figure}
In Fig.\,\ref{fig:regularizations}, we compare the test error of an unregularized student learning from the true labels (grey curve) with the test error obtained by learning from the outputs of an $L_2$-regularized teacher (with different intensity levels, blue and red curves). We call this a ``pure distillation'' setting, with $\chi=1$ and $\lambda=0$. In black we show the Bayes optimal lower bounds for teacher and student. The corresponding performance of the teacher is displayed with dashed lines. 

The first observation we can make is that, indeed, there is a transfer of the regularization properties from the teacher to the student, since the soft outputs of the teacher inform the student against overfitting the training set and growing the norm of the weights disproportionately. The second observation is that better teacher regularization induces better student regularization. In particular, we can see that in the large $\alpha$ regime, a better student performance is attained at the value of $\tilde \lambda$ which also optimizes the teacher performance. The student is thus able to inherit the fine-tuning done at the level of the teacher, even though it does not belong to the same model class. Note however that the KD student test error is still far from the displayed Bayes optimal bound.

In this plot we avoided showing the generalization behavior in the low $\alpha$ regime, which will be described in detail in section \ref{sec:double_descent}. Moreover, a more thorough analysis of the location of the optimal teacher regularization and the corresponding student performance can be found in Appendix\,\ref{sec:more_on_inheritance}.

\subsection{Limits of KD}

Now that we have seen that KD can indirectly regularize the student, we continue by studying whether it can outperform direct regularization methods. 

First, we compare the generalization curves obtained in the pure distillation setting described above ($\chi=1$, learning only from the teacher outputs) with the effect of a simple $L_2$ penalty directly at the level of the student loss ($\chi=0$, learning only from the true labels).
\begin{figure}[ht]
\centering
\includegraphics[width=0.5\textwidth]{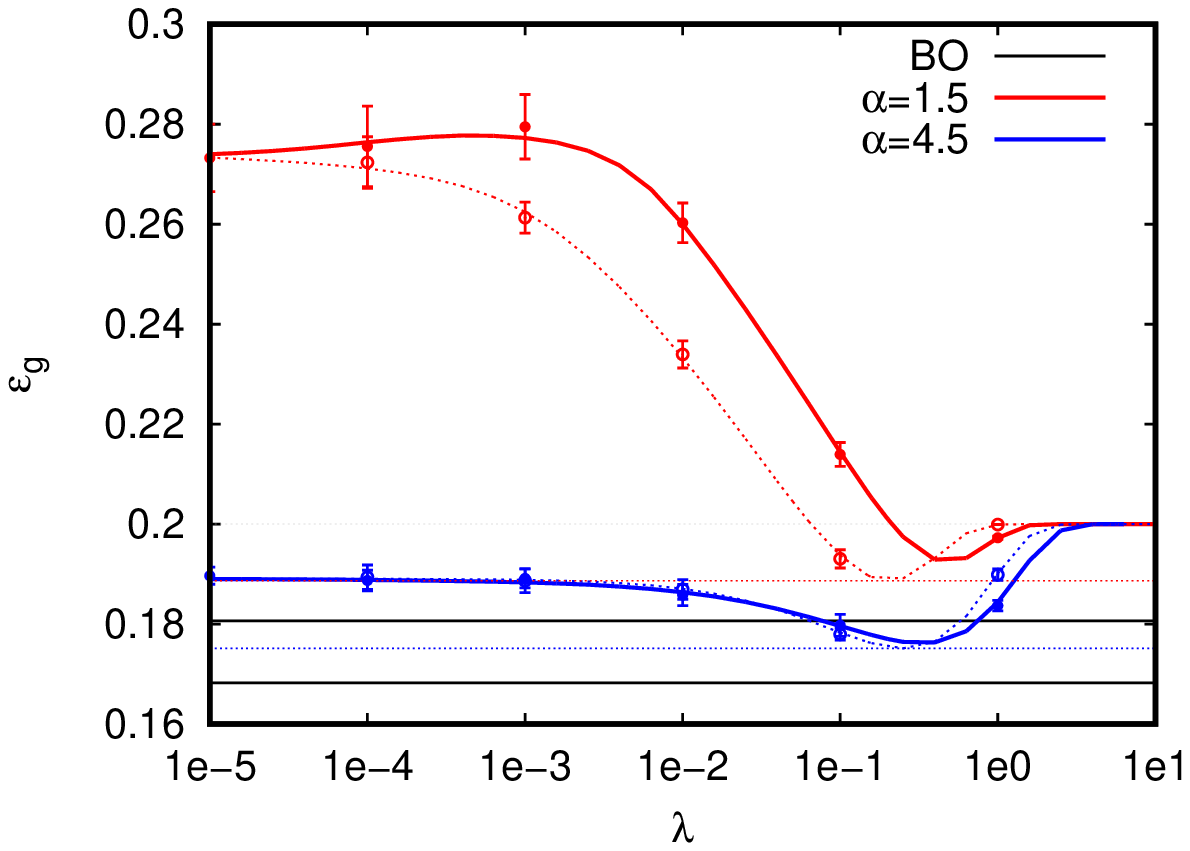}
\includegraphics[width=0.5\textwidth]{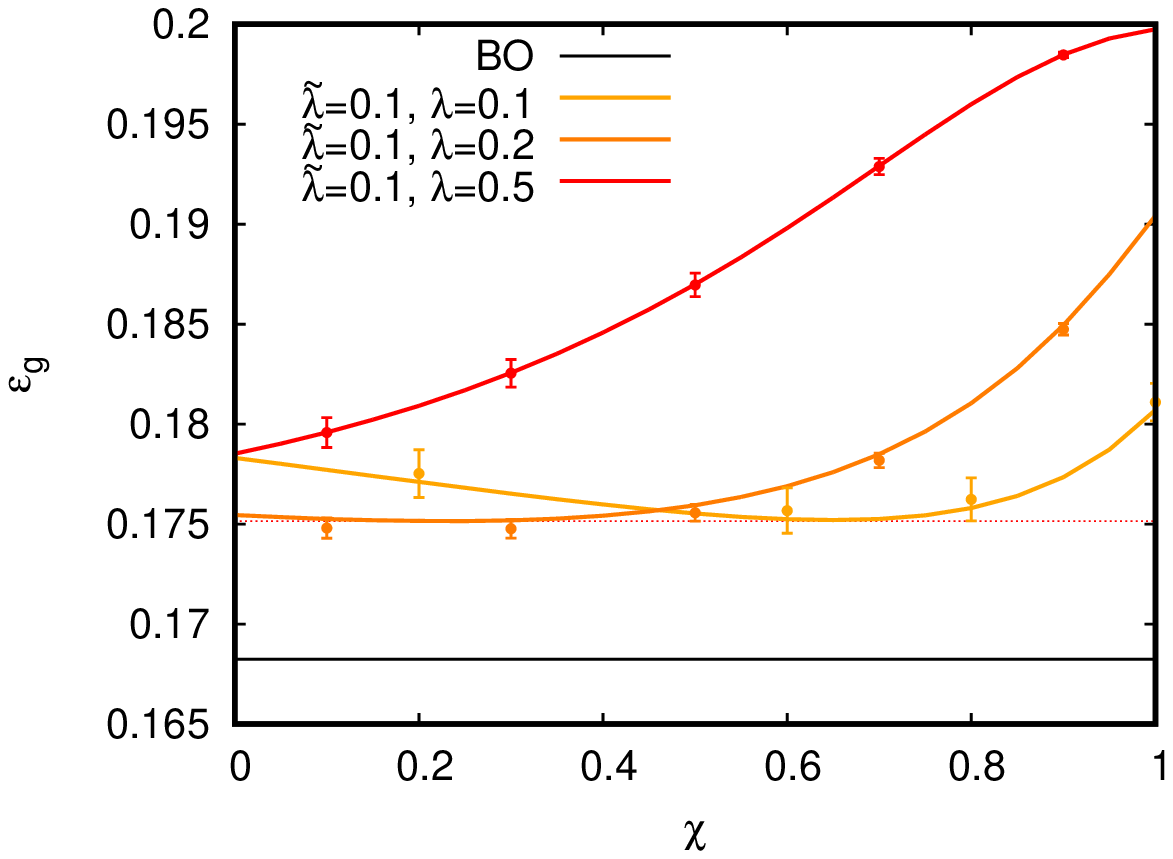}
\caption{\label{fig:dist_vs_reg} \emph{Top plot}: Comparison between the test error of a student learning from a regularized teacher (full curves) with $\chi=1$, and a directly regularized student (dashed curves) with $\chi=0$, at fixed values of $\alpha$ and at $\rho=0.2$, $\Delta=1$, $T=1$ and $\eta=0.5$. \emph{Bottom plot}: variation in the test error induced by tuning the mixing parameter $\chi$, at $\alpha=4.5$, in correspondence of three regularization regimes in the student loss (under-regularized, properly regularized, over-regularized). (Horizontal dashed lines) Test error achieved by setting the optimal $L_2$ regularization intensity (at $\chi=0$). (Black lines) Bayes optimal performance. The data points with error bars represent the results of numerical experiments at $N=4000$ ($10$ samples per point).}
\end{figure}
In the top plot of Fig.\,\ref{fig:dist_vs_reg}, we display the test error in the two cases (full lines for $\chi=1$, dashed lines for $\chi=0$) at two dataset sizes $\alpha=1.5$ (red) and $\alpha=4.5$ (blue). The horizontal dashed lines highlight the best direct regularization performance. The black horizontal lines, instead, show the Bayes optimal bound (top $\alpha=1.5$, bottom $\alpha=4.5$). One can see that the two curves differ the most in the low $\alpha$ regime, where the direct regularization is outperforming KD  (this will be clarified in Sec.\,\ref{sec:double_descent}). At higher values of $\alpha$, instead, the effects of regularizing the teacher or directly regularizing the student and the associated generalization performances become almost indistinguishable. This is a positive result: with no fine-tuning at the level of the student loss a better teachers does induce a better student performance. However, it is clear that in this case pure distillation is not leading to any improvements over simple ridge regularization.

We thus consider a second setting, where we turn on the student $L_2$ regularization intensity $\lambda$ also in the KD loss and then vary the mixing parameter $\chi$ in order to balance the total amount of regularization. In the bottom plot of Fig.\,\ref{fig:dist_vs_reg}, we fix $\alpha=4.5$ and $\tilde{\lambda}=0.1$ (optimal regularization regime for the teacher) and explore three values of $\lambda$ for the student: over-regularized case ($\lambda=0.5$), properly regularized case ($\lambda=0.2$) and under-regularized case ($\lambda=0.1$). Again, the dashed horizontal line and the black horizontal line respectively mark the best performance achieved with perfect fine-tuning of the ridge regularization (at $\chi=0$) and the Bayes optimal performance. We observe the following: in the over-regularized regime, adding even more regularization through KD is not beneficial and the best value of the mixing parameter is thus $\chi=0$ (the minimum of the generalization is sub-optimal with respect to the grey line); in the other settings, instead, one finds an optimal value of $\chi$ at which the performance associated with optimal direct regularization is matched. 

Overall these two experiments show a clear limitation of KD in the studied model: its effect is at best equal to that of a fine-tuned direct regularization scheme, and thus the student performance is still inferior with respect to the Bayes optimal one. In Appendix\,\ref{sec:more_on_inheritance} we provide further details and take into account also a different setting, where the regularization scheme is based on the introduction of soft-labels: we can report here that also in that case the direct regularization and the inherited KD regularization induce the same generalization performance.
It, of course, remains to be seen whether this above limitation of KD extends beyond the studied model.

\subsection{Learning from a Bayes Optimal teacher} 

We finally consider a case where the teacher is not trained through an explicit regularization method: since in this setting it is not possible to regularize directly the student in a similar way, a transfer learning strategy becomes necessary. This construction is meant to mimic more closely the behavior of knowledge distillation in usual deep learning settings, where learning algorithms play an implicit role in regularizing the network and their effectiveness may be dependent on the architecture.

In our framework, we study the performance of a student distilling the knowledge of a Bayes optimal teacher. As mentioned before, in the GM model there exists a point estimator that achieves the Bayes optimal generalization performance Eq.\,(\ref{eq:plug-in}), which is characterized by an overlap with the signal $v$, a norm and a bias respectively equal to:
\begin{equation}
\tilde{m} = \frac{\boldsymbol{v}\cdot \boldsymbol{w}_{BO}}{N}=1, \,\,\,\,\,\,\ \tilde{q}=\frac{\lVert\boldsymbol {w}_{BO} \rVert^2}{N}=1+\Delta/\alpha , \,\,\,\,\,\,\, 
\tilde{b} = \frac{\Delta(1+\Delta/\alpha)}{2}\log(\frac{\rho}{1-\rho}).
\end{equation}
\begin{figure}[ht]
\centering
\includegraphics[width=0.5\textwidth]{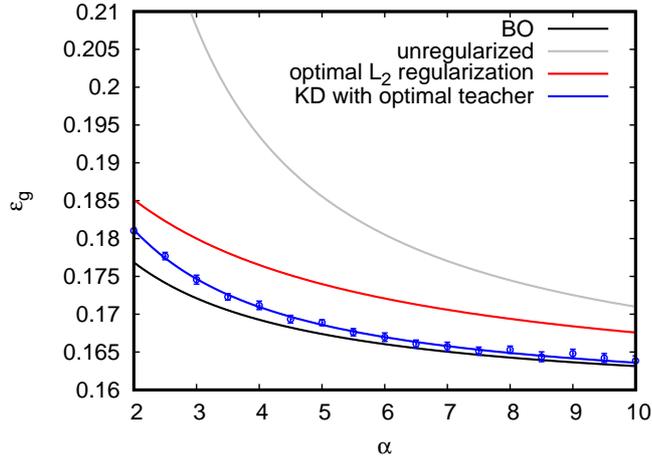}
\caption{\label{fig:BO_teach} Generalization performance of a $\eta=0.5$ sparse student in a GM setting ($\rho=0.2$, $\Delta=1$). (Black curve) Performance bound, given by the generalization of the sparsified plug-in estimator. (Grey curve) Unregularized case. (Red curve) Student learning with a direct $L_2$ regularization of optimal intensity. (Blue curve) Pure distillation student ($\chi=1$) learning from an optimal teacher (e.g. BO point estimator). The data points with error bars represent the results of numerical experiments at $N=4000$ ($10$ samples per point).
}
\end{figure}
In order to obtain an analytical characterization of this special distillation setting, instead of taking the Hebbian estimator of Eq.\,(\ref{eq:plug-in}) directly (carrying non-trivial correlations with the noise terms in the training data-points), we consider a teacher with the same bias and weights distributed as: 
\begin{equation}
\tilde{\boldsymbol{w}} = \boldsymbol{v} + \sqrt{\frac{\Delta}{\alpha}}\boldsymbol{z},
\end{equation}
where each component of $\boldsymbol{z}$ is i.i.d. Gaussian distributed with unit variance. A typical realization of this teacher will achieve exactly the Bayes optimal generalization performance, so we can use it as a proxy for studying the distillation setting with the Hebbian estimator as a teacher. The details of the associated analytical calculation are reported in Appendix\,\ref{sec:replica_computations}.

In Fig.\,\ref{fig:BO_teach} we compare the generalization performance in three different learning settings: (grey line) an unregularized student, (red line) a student with direct $L_2$ regularization set at the optimal intensity and learning from the ground truth labels, and (blue line) a pure distillation student ($\chi=1$) learning from a Bayes optimal teacher (according to the above definition). These results are again compared with the Bayes optimal performance bound (black line) relevant for the student. 

When the transferred outputs are produced by an optimal teacher, we observe a clear improvement in the distillation test error with respect to a direct $L_2$ regularization: the gap with the optimal generalization bound is nearly closed (especially at large values of $\alpha$). This result clearly shows the potential of knowledge distillation: through transfer learning the student can reach performances that are seemingly not achievable with direct regularization schemes, inheriting also the``implicit'' regularization of the teacher (similar to what is observed in deep learning experiments \cite{frankle2018lottery}).

\section{Double descent in the KD framework}
\label{sec:double_descent}

In this final section we will focus on the low $\alpha$ regime, i.e. number of samples small or comparable to the dimensionality, and the limit of zero direct regularization either in the student or in the teacher losses. 

Note that, in the considered model, when the $L_2$ regularization term is completely switched off, the minimization of non-regularized logistic loss becomes equivalent to maximum likelihood estimation (MLE). In the GMM, with high probability the generated binary data will be linearly separable up until some threshold $\alpha_S(\rho,\Delta,\eta)$, and in this regime the ML estimator is ill-defined \cite{sur2019modern}. Since we are not interested in addressing this issue in the present work, in the following we will consider a baseline regularization intensity of $\lambda=0.00001$ as a proxy for the unregularized limit.

Let's first consider taking an unregularized teacher and a pure distillation student ($\chi=0$, $\tilde{\lambda},\lambda \to 0$). The teacher training problem is perfectly separable below the separability threshold $\tilde{\alpha}_S$: without an explicit regularization its norm will thus diverge, due to the shape of the cross-entropy loss. Therefore, the outputs produced by the teacher will be quasi-binary (since the sigmoid activation will be completely saturated), and the student learning problem will look exactly like the usual logistic regression with binary labels. If we focus on the generalization behavior of the student, we thus expect a peak (and the corresponding double-descent behavior) at the student's linear separability threshold $\alpha_S$ \cite{mignacco2020role}. Note that in general $\alpha_S\neq\tilde{\alpha}_S$, since we take $\eta<1$.

Now, let's consider instead the case of a regularized teacher. In this case, even below $\tilde{\alpha}_S$, the teacher norm will remain finite and the produced outputs will be continuously distributed in the range $[0,1]$. In this case, the pure distillation student will try to interpolate a set of non-binary outputs: as long as the number of linear constraints will be lower than the number of trainable weights $\alpha<\eta$, the student will be able to exactly reproduce the teacher outputs. However, just like in a normal regression scenario \cite{mignacco2020role}, this will give rise to an interpolation peak at $\alpha = \eta$ (and the associated double-descent).

\begin{figure}[ht]
\centering
\includegraphics[width=0.5\textwidth]{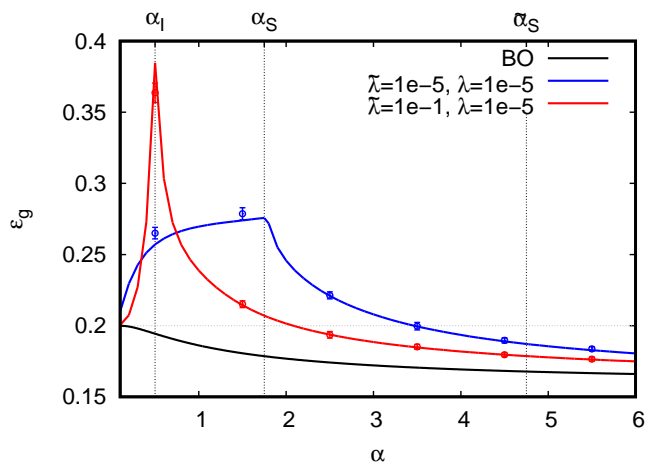}
\caption{\label{fig:peaks} Generalization performance of a $\eta=0.5$ sparse student in a pure distillation setting ($\rho=0.2$, $\Delta=1$, $\chi=1$). (Black curve) Performance bound, given by the Bayes Optimal generalization. (Blue curve) Distillation curve when both teacher and student are not $L_2$ regularized. (Red curve) Distillation performance when only the teacher is regularized. $\alpha_I$, $\tilde{\alpha}_S$ and $\alpha_S$ respectively denote the student interpolation threshold and the teacher and student separability thresholds. The data points with error bars represent the results of numerical experiments at $N=4000$ ($10$ samples per point).
}
\end{figure}
In Fig.\,\ref{fig:peaks}, we can see the two types of peak (blue and red curves respectively). Note that the deviation of the experimental points from the theoretical predictions, at low $\alpha$ in the first regime, is due to the non convergence of the gradient descent simulations before the cutoff of $2000$ optimization epochs. 

In Appendix\,\ref{sec:more_on_double-descent}, we also provide an in-depth analysis of teacher and student training losses and the mean squared error between teacher and student activations/preactivations around $\alpha_I$ and $\alpha_S$, in order to further clarify the described phenomena.

\section{Conclusions}\label{sec:conclusions}

In this work we developed a statistical physics framework for analysing knowledge distillation in high-dimensional models solvable with the replica method. The framework yields a deterministic description of the typical properties of the studied model, via a set of fixed point equations that track the behavior of the relevant order parameters.

We applied our framework to a prototypical case of knowledge distillation in the presence of mismatch between teacher and student networks. In particular, we considered two linear classifiers with different support (a stronger teacher network and a weaker student network) trained over a binary classification problem with data generated according to a Gaussian mixture. 

We were able to highlight the inheritance properties of KD, showing that learning from properly regularized teachers can effectively transfer the good generalization properties to the student, with little fine-tuning at the level of the distillation loss. In our model, we also showed that, by distilling the knowledge of a Bayes optimal teacher, an agnostic KD student can approach the optimal generalization bound, whereas usual regularized logistic regression remains clearly sub-optimal. In order to validate the theoretical predictions, we offered a comparison with the results of finite size numerical experiments.

Finally, we analyzed the peculiar double-descent phenomenology that can appear in a KD setting, hybridizing between the interpolation peaks in regression problems and the separability threshold peak in classification problems.

The presented analytic results were obtained through the non-rigorous (yet exact) replica method, but the convex nature of the studied optimization problems suggests the possibility of an independent rigorous derivation via the Gordon minimax theorem \cite{gordon1985some}, along the lines of \cite{mignacco2020role}. We leave this technical goal for future work.

Another natural but challenging research direction is to consider the case of more realistic data generative models (e.g, \cite{goldt2019modelling, gerace2020generalisation}) and, more importantly, more complex neural network architectures (e.g., a random features \cite{mei2019generalization}, or one hidden-layer networks \cite{aubin2018committee}) that could allow for a more realistic mismatch between teacher and student models.

\acknowledgements{
We thank St\'ephane d'Ascoli for his inspiring presentation of the knowledge distillation problem. We acknowledge funding from the ERC under the European Union’s
Horizon 2020 Research and Innovation Programme Grant Agreement 714608-SMiLe.

}

\bibliography{bibliography}
\newpage

\appendix

\section{Replica Computations}
\label{sec:replica_computations}

\subsection{Typical learning in the GMM}

We are interested in evaluating the following average free-entropy:
\begin{equation}
\Phi=\lim_{\beta\to\infty}\frac{1}{\beta N}\left\langle \log\int d\boldsymbol{w} e^{-\frac{\beta\lambda}{2}\lVert \boldsymbol{w} \rVert^2}\int db\prod_{\mu}e^{-\beta\,\ell\left(y^{\mu},\sigma\left(\sum_{i=1}^{N}\frac{w_{i}x_{i}^{\mu}}{\sqrt{N}}+b\right)\right)}\right\rangle _{\left\{ \boldsymbol{x}^{\mu},y^{\mu}\right\} } \label{eq:typ_FE}
\end{equation}
where $\ell(\cdot)$ is the cross-entropy loss and the training data $\{\boldsymbol{x}^\mu,y^\mu{}{}\}$ is distributed according to a Gaussian Mixture model:
\begin{equation}
    \boldsymbol{x}^{\mu}=\left(2y^{\mu}-1\right)\frac{\boldsymbol{v}}{\sqrt{N}}+\boldsymbol{z}
\end{equation}
with $v_{i},z_{i}\sim\mathcal{N}\left(0,\Delta\right)$ and $y^{\mu}\sim\rho\,\delta\left(y^{\mu}-1\right)+(1-\rho)\,\delta\left(y^{\mu}\right)$.
Note that, because of the isotropy of the data model, instead of integrating over the possible realizations for the signal vector $\boldsymbol{v}$ it is possible to fix
the gauge $\boldsymbol{v}=(1,1,...,1)^{T}$ (e.g., as in \cite{engel2001statistical}).

In the $\beta\to\infty$ limit, the statistical measure on the weights and the bias $\{\boldsymbol{w},b\}$ focuses over the minimizer of the training loss, which yields the logistic regression optimization problem we want to characterize through this calculation.
In order to evaluate the quenched average appearing in Eq.\,(\ref{eq:typ_FE}), we resort to the non-rigorous Replica Method, introduced in the context of Disordered Systems \cite{mezard1987spin} and based on the identity:
\begin{equation}
    \log(x) = \lim_{n\to0} \frac{x^n-1}{n}.
\end{equation}
So, instead of evaluating the average of the free-energy directly, we introduce $n$ interacting replicas of the original system and in the end we will recover the original expression by extrapolating the limit $n\to0$. 

We can thus focus on the calculation of the replicated volume:
\begin{equation}
\Omega^{n}=\int\prod_{a}d\boldsymbol{w}^{a}e^{-\frac{\beta\lambda}{2}\lVert \boldsymbol{w} \rVert^2}\int\prod db^{a}\prod_{\mu}\prod_{a}\left\langle e^{-\beta\,\ell\left(y^{\mu},\sigma\left(\sum_{i=1}^{N}\frac{w_{i}^{a}x_{i}^{\mu}}{\sqrt{N}}+b^{a}\right)\right)}\right\rangle _{\left\{ \boldsymbol{x}^{\mu},y^{\mu}\right\} }
\end{equation}
\begin{equation}
=\int\prod_{a}d\boldsymbol{w}^{a}e^{-\frac{\beta\lambda}{2}\lVert \boldsymbol{w} \rVert^2}\int\prod_{a}db^{a}\int\prod_{\mu}\prod_{a}\frac{d\lambda_{\mu}^{a}d\hat{\lambda}_{\mu}^{a}}{2\pi}e^{i\hat{\lambda}_{\mu}^{a}\left(\lambda_{\mu}^{a}-\sum_{i=1}^{N}\frac{w_{i}^{a}x_{i}^{\mu}}{\sqrt{N}}\right)} \times, \nonumber
\end{equation}
\begin{equation}
    \prod_{\mu}\prod_{a}\left\langle e^{-\beta\,\ell\left(y^{\mu},\sigma\left(\lambda_{\mu}^{a}+b^{a}\right)\right)}\right\rangle _{\left\{ \boldsymbol{x}^{\mu},y^{\mu}\right\}}
\end{equation}
where in the second line we isolated the dependency on each training data point $\boldsymbol{x}^\mu,y^\mu$ by introducing the preactivation variables $\lambda^a_\mu$ via Dirac's $\delta$ functions, allowing us to take the disorder average:
\begin{equation}
\mathbb{E}_{\boldsymbol{x}^{\mu}}e^{-i\sum_{a}\hat{\lambda}_{a}^{\mu}\frac{\boldsymbol{x}^{\mu}\cdot\boldsymbol{w}^{a}}{\sqrt{N}}} = e^{-i\left(2y^{\mu}-1\right)\sum_{a}\hat{\lambda}_{a}^{\mu}\frac{\sum w_{i}^{a}v_{i}}{N}}e^{-\frac{\Delta}{2}\sum_{ab}\hat{\lambda}_{a}^{\mu}\hat{\lambda}_{b}^{\mu}\frac{\sum_{i}w_{i}^{a}w_{i}^{b}}{N}} + \mathcal{O}(N^{-3/2 }).
\end{equation}
Now, we can introduce the overlap order parameters:
\begin{itemize}
\item $m^{a}=\frac{\sum w_{i}^{a}v_{i}}{N}$, representing the magnetization, i.e. the overlap between the learned weight configuration and the true signal $v$.
\item $q^{ab}=\frac{\sum_{i}w_{i}^{a}w_{i}^{b}}{N}$, representing the overlap between two configurations sampled from the measure Eq.\,(\ref{eq:typ_FE}).
\end{itemize}
Then, we can rewrite the replicated volume as:
\begin{equation}
\Omega^{n}=\int\prod_{a}\frac{dm^{a}d\hat{m}^{a}}{2\pi/N}\int\prod_{ab}\frac{dq^{ab}d\hat{q}^{ab}}{2\pi/N}\int\prod db^{a}G_{I}\left(G_{S}\right)^{N}\left(G_{E}\right)^{\alpha N}
\end{equation}
where we separated the action in three different contributions: an interaction term, containing a trace over the order parameters and their conjugates
\begin{equation}
G_{I}=\exp\left(-N\left(\sum_{a}\hat{m}^{a}m^{a}+\sum_{ab}\hat{q}^{ab}q^{ab}\right)\right),
\end{equation}
an entropic term, factorized over the components of the weight vector and containing the information about the regularization term in the loss function
\begin{equation}
G_{S}=\int\prod_{a}d w^{a} e^{-\frac{\beta\lambda}{2} ({w^a})^2}\exp\left(\sum_{a}\hat{m}^{a}w^{a}+\sum_{ab}\hat{q}^{ab}w^{a}w^{b}\right),
\end{equation}
and an energetic term, factorized over the training data points and containing the cross-entropy term:
\begin{equation}
G_{E}=\int\prod_{a}\left(\frac{d\lambda^{a}d\hat{\lambda}^{a}}{2\pi}e^{i\lambda^{a}\hat{\lambda}^{a}}\right)e^{-\frac{\Delta}{2}\sum_{ab}\hat{\lambda}_{a}\hat{\lambda}_{b}q^{ab}}\left\langle \prod_{a}e^{-\beta\,\ell\left(y,\sigma\left(\lambda^{a}+(2y-1)m^{a}+b^{a}\right)\right)}\right\rangle _{y}.
\end{equation}

\subsubsection{Replica symmetric Ansatz}
In order to proceed in the computation, we have to make an assumption on the structure of the order parameters and the geometric organization of the $n$ replicas of the original system. Because of the convexity of the optimization problem we are considering we are justified in adopting the simplest possible assumption, the so-called Replica Symmetric ansatz, posing:
\begin{itemize}
\item $m^{a}=m$ for all $a=1,...,n$, and same for their conjugates.
\item $q^{ab}=q$ for all $a>b$, $q^{ab}=Q$ for all $a=b$, and same for their
conjugates.
\item $b^{a}=b$ for all $a=1,...,n$.
\end{itemize}

Then, one can substitute the RS ansatz in the interaction term and easily obtain:
\begin{equation}
\frac{\log G_{I}}{nN}=g_I
=-\left(\hat{m}m+\frac{\hat{Q}Q}{2}-\frac{\hat{q}q}{2}\right)
\end{equation}

In the entropic term, after the substitutions one gets:
\begin{equation}
G_{S}=\int\prod_{a}d w^{a}  e^{-\frac{\beta\lambda}{2} ({w^a})^2} \exp\left(\hat{m}\sum_{a}w^{a}+\frac{1}{2}\left(\hat{Q}-\hat{q}\right)\sum_{a}\left(w^{a}\right)^{2}+\frac{1}{2}\hat{q}\left(\sum_{a}w^{a}\right)^2\right)
\end{equation}
\begin{equation}
=\int\mathcal{D}z_{0}\left\{ \int d w  e^{-\frac{\beta\lambda}{2} ({w})^2} \exp\left(\frac{1}{2}\left(\hat{Q}-\hat{q}\right)w^{2}+\left(\hat{m}+\sqrt{\hat{q}}z_{0}\right)w\right)\right\} ^{n}
\end{equation}
where in the second line a Hubbard-Stratonovich transformation, introducing the auxiliary variable $z_0\sim\mathcal{N}(0,1)$, allowed factorization over the replica index. Now we can take the logarithm in the $n\to0$ limit, obtaining:
\begin{equation}
\frac{\log G_S}{nN} = g_{S} = \int\mathcal{D}z_{0}\log\int d w  \exp\left(\frac{1}{2}\left(\hat{Q}-\hat{q}-\beta\lambda\right)w^{2}+\left(\hat{m}+\sqrt{\hat{q}}z_{0}\right)w\right).
\end{equation}
Similarly, one can obtain the RS energetic contribution:
\begin{equation}
G_E=\mathbb{E}_y\int\prod_{a}\left(\frac{d\lambda^{a}d\hat{\lambda}^{a}}{2\pi}e^{i\lambda^{a}\hat{\lambda}^{a}}\right)e^{-\frac{\Delta}{2}\left(Q-q\right)\sum_{a}\left(\hat{\lambda}_{a}\right)^{2}-\frac{\Delta}{2}q\left(\sum_{a}\hat{\lambda}_{a}\right)^{2}} \prod_{a}e^{-\beta\,\ell\left(y,\sigma\left(\lambda^{a}+(2y-1)m+b\right)\right)}
\end{equation}
\begin{equation}
=\mathbb{E}_y\int\mathcal{D}z_{0}\left\{ \int\frac{d\lambda}{\sqrt{2\pi}}\frac{1}{\sqrt{\Delta\left(Q-q\right)}}e^{-\frac{1}{2}\frac{\left(\lambda+\sqrt{\Delta q}z_{0}\right)^{2}}{\Delta\left(Q-q\right)}}e^{-\beta\,\ell\left(y,\sigma\left(\lambda+(2y-1)m+b\right)\right)}\right\} ^{n}
\end{equation}
\begin{equation}
=\mathbb{E}_y\int\mathcal{D}z_{0} \left\{ \int\mathcal{D}\lambda e^{-\beta\,\ell\left(y,\sigma\left(\sqrt{\Delta\left(Q-q\right)}\lambda+\sqrt{\Delta q}z_{0}+(2y-1)m+b\right)\right)}\right\} ^{n}
\end{equation}
so taking the logarithm:
\begin{equation}
\frac{\log{G_{E}}}{n}=g_{E}=\mathbb{E}_y\int\mathcal{D}z_{0} \log \int\mathcal{D}\lambda e^{-\beta\,\ell\left(y,\sigma\left(\sqrt{\Delta\left(Q-q\right)}\lambda+\sqrt{\Delta q}z_{0}+(2y-1)m+b\right)\right)}.
\end{equation}

\subsubsection{Zero temperature limit}
Now we can focus on the $\beta\to\infty$ limit, in which we recover the origin empirical risk minimization problem. Because of convexity, as we lower the temperature the overlap $q$ between two different configurations sampled from the statistical measure will approach the typical norm $Q$, suggesting the scaling:
\begin{equation}
\left(Q-q\right)=\delta q/\beta
\end{equation}
Moreover, the conjugate parameters also need to be properly rescaled:
\begin{equation}
\left(\hat{Q}-\hat{q}\right)=-\beta\delta\hat{q}, \,\,\,\,\,
\hat{q}\sim\beta^{2}\hat{q}, \,\,\,\,\,
\hat{m}\sim\beta\hat{m}.
\end{equation}
which gives, for the interaction term:
\begin{equation}
g_{i}=-\beta\left(\hat{m}m+\frac{1}{2}\left(\hat{q}\delta q-\delta\hat{q}q\right)\right)
\end{equation}

In the entropic term one gets an integration over a one-dimensional optimization problem:
\begin{equation}
g_{s}=\beta\int\mathcal{D}z_{0}\max_{w}\left(-\frac{\lambda+\delta\hat{q}}{2}w^{2}+\left(\hat{m}+\sqrt{\hat{q}}z_{0}\right)w\right)
\end{equation}
where the maximum is located at:
\begin{equation}
w^{\star}=\frac{\left(\hat{m}+\sqrt{\hat{q}}z_{0}\right)}{\left(\lambda+\delta\hat{q}\right)} \label{eq:teach_max}
\end{equation}
and the expresssion can be evaluated analytically, giving:
\begin{equation}
g_{s}=\beta\frac{\hat{m}^{2}+\hat{q}}{2\left(\lambda+\delta\hat{q}\right)}.
\end{equation}

Finally, in the energetic contribution we get:
\begin{equation}
g_{E}=\mathbb{E}_y \int\mathcal{D}z M_{E}
\end{equation}
\begin{equation}
M_{E}=\max_{u}-\frac{u^{2}}{2}-\ell\left(y,\sigma\left(\sqrt{\Delta\delta q}u+\sqrt{\Delta q}z+\left(2y-1\right)\,m+b\right)\right)
\end{equation}

and one obtains the free-entropy:
\begin{equation}
\Phi=-\left(\hat{m}m+\frac{1}{2}\left(\hat{q}\delta q-\delta\hat{q}q\right)\right)+g_{S}+\alpha g_{E}. \label{eq:T-FE}
\end{equation}

The fixed-point equation that characterize the logistic regression problem in the high dimensional limit are obtained by extremizing the free-entropy with respect to the order parameters and their conjugates, which is nothing but a saddle-point condition for the action in the Statistical Physics framework.

\begin{figure}[ht]
\centering
\includegraphics[width=0.5\textwidth]{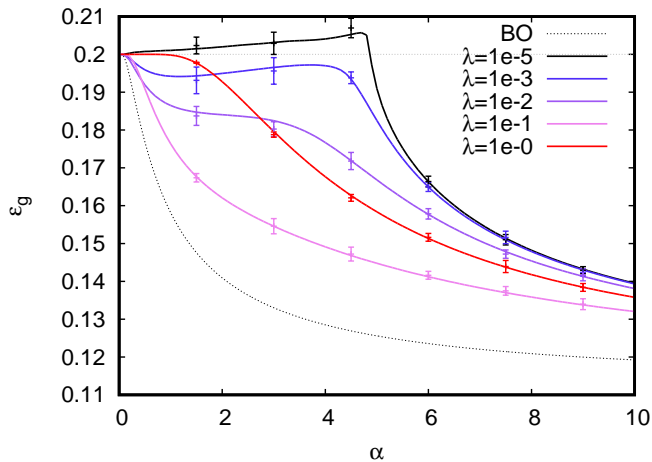}
\caption{\label{fig:typ_learning} Classification in the GM model with $\rho=0.2$ and $\Delta=1$. Colored lines: linear classifier trained with logistic regression and ridge regularization. Dashed black line: plug-in estimator achieving the Bayes optimal performance.}
\end{figure}

In Fig.\,\ref{fig:typ_learning} we compare the generalization performance for a model trained with $L_2$-regularized logistic regression with the Bayes-optimal performance, in the $\rho=0.2$, $\Delta=1$ setting. This will represent the baseline teacher model in the study of the distillation process. Note that the optimal value for the regularization parameter is of order $\lambda\sim 1e-1$ and that higher values will hinder the generalization performance.

\subsection{Distillation in the GMM}

We will now present the derivation of the free energy expression for the distillation framework analyzed in the main text. In the following, all the parameters that refer to the teacher linear classifier will be denoted with an additional tilde.

As mentioned, in order to avoid the trivial scenario where the student converges exactly to the weight configuration of the teacher, we will consider a weaker ``student''
model, with a $0<\eta<1$ fraction of weights set to zero throughout the learning process. Thus, we aim to evaluate the expected value (in high dimensions) of the following free-entropy:
\begin{equation}
\Phi=\lim_{\beta\to\infty}\frac{1}{\beta N}\left\langle \left\langle \log\int \prod_{i=1}^{\eta N} d w_i e^{-\frac{\beta\lambda}{2}\lVert \boldsymbol{w} \rVert^2}\int db\prod_{\mu}e^{-\frac{\beta}{2}\ell\left(\sigma\left(\frac{\tilde{\boldsymbol{w}}\cdot \boldsymbol{x}^{\mu}}{\sqrt{N}}+\tilde{b}\right),\sigma\left(\frac{\boldsymbol{w}\cdot \boldsymbol{x}^{\mu}}{\sqrt{N}}+b\right)\right)}\right\rangle _{\tilde{\boldsymbol{w}}}\right\rangle _{\left\{ \boldsymbol{x}^{\mu},y^{\mu}\right\} } \label{eq:dist_FE}
\end{equation}
where the internal brackets represent an average over the teacher $\{\tilde{\boldsymbol{w}},\tilde{b}\}$ measure. 

In order to evaluate the internal average we will use a different version of the replica trick, based on the following identity: 
\begin{equation}
    \left\langle f(\tilde{\boldsymbol{w}}) \right\rangle_{\tilde{\boldsymbol{w}}} = \frac{\int d\mu(\tilde{\boldsymbol{w}}) f(\boldsymbol{w})} {\int d\mu(\tilde{\boldsymbol{w}})} =
    \lim_{\tilde{n}\to 0} \int \prod_{c=0}^{\tilde{n}} d\mu(\tilde{\boldsymbol{w}}^c) f(\tilde{\boldsymbol{w}}^1)
\end{equation}
introducing $\tilde{n}$ replicas of the teacher configuration but coupling only the first one to the student system. In this way, in the $\tilde{n}\to 0$ limit we can recover 
the expectation over the teacher measure. This type of formalism is closely related to the seminal work \cite{franz1997phase} and was more recently applied for example in \cite{huang2013entropy,baldassi2016learning,baldassi2018role}.

As in the previous computation, the quenched disorder average can be taken only by replacing the logarithm in the definition of the free-entropy with the $n\to0$ limit of the replicated system, so we will focus on the evaluation of:
\begin{equation}
\frac{1}{N}\lim_{n,\tilde{n}\to0}\partial_{n}\left<\lim_{\tilde{\beta}\to\infty}\lim_{\beta\to\infty}\int\prod_{c=1}^{\tilde{n}}d \tilde{\boldsymbol{w}}^{c} e^{-\frac{\beta\tilde{\lambda}}{2}\lVert \tilde{\boldsymbol{w}}^c \rVert^2} \int\prod_{c=1}^{\tilde{n}}d\tilde{b}^{c}\prod_{\mu,c}e^{-\frac{\tilde{\beta}}{2}\ell\left(y^{\mu},\sigma\left(\sum_{i=1}^{N}\frac{\tilde{w}_{i}^{c}x_{i}^{\mu}}{\sqrt{N}}+\tilde{b}^{c}\right)\right)}\right. \nonumber
\end{equation}
\begin{equation}
\left.\times\int\prod_{a=1}^{n}d \boldsymbol{w}^{a} e^{-\frac{\beta\lambda}{2}\lVert \boldsymbol{w}^a \rVert^2} \int\prod_{a=1}^{n}db^{a}\prod_{\mu,a}e^{-\frac{\beta}{2}\ell\left(\sigma\left(\sum_{i=1}^{N}\frac{\tilde{w}_{i}^{1}x_{i}^{\mu}}{\sqrt{N}}+\tilde{b}^{1}\right),\sigma\left(\sum_{i=1}^{N}\frac{w_{i}^{a}x_{i}^{\mu}}{\sqrt{N}}+b^{a}\right)\right)}\right>_{\left\{ \boldsymbol{x}^{\mu},y^{\mu}\right\} }.
\end{equation}
Following the same steps as above, we introduce the Dirac's $\delta$ (integral representation) for teacher and student preactivations, but in this case we will also separate the first $\eta$ components  of the teacher (where the student weights are non-zero) from the rest:
\begin{equation}
1 = \int\prod_{\mu,c}\frac{d\tilde{u}_{\mu}^{c}d\hat{\tilde{u}}_{\mu}^{c}}{2\pi}e^{i\hat{\tilde{u}}_{\mu}^{c}\left(\tilde{u}_{\mu}^{c}-\sum_{i=\eta N+1}^{N}\frac{\tilde{w}_{i}^{c}x_{i}^{\mu}}{\sqrt{N}}\right)}\int\prod_{\mu,c}\frac{d\tilde{\lambda}_{\mu}^{c}d\hat{\tilde{\lambda}}_{\mu}^{c}}{2\pi}e^{i\hat{\tilde{\lambda}}_{\mu}^{c}\left(\lambda_{\mu}^{c}-\sum_{i=1}^{\eta N}\frac{\tilde{w}_{i}^{c}x_{i}^{\mu}}{\sqrt{N}}\right)}\times \nonumber
\end{equation}
\begin{equation}
    \int\prod_{\mu,a}\frac{d\lambda_{\mu}^{a}d\hat{\lambda}_{\mu}^{a}}{2\pi}e^{i\hat{\lambda}_{\mu}^{a}\left(\lambda_{\mu}^{a}-\sum_{i=1}^{\eta N}\frac{w_{i}^{a}x_{i}^{\mu}}{\sqrt{N}}\right)}.
\end{equation}
Now we separately take the disorder average over the non-zeros components:
 \begin{eqnarray}
 & \prod_{i=1}^{\eta N}\mathbb{E}_{x_{i}^{\mu}}e^{-i\left(\sum_{c}\hat{\tilde{\lambda}}_{c}^{\mu}\frac{\tilde{w}_{i}^{c}}{\sqrt{N}}+\sum_{a}\hat{\lambda}_{a}^{\mu}\frac{w_{i}^{a}}{\sqrt{N}}\right)x_{i}^{\mu}} = \\
&= e^{-i\left(2y^{\mu}-1\right)\left(\sum_{c}\hat{\tilde{\lambda}}_{c}^{\mu}\frac{\sum_{i=1}^{\eta N}\tilde{w}_{i}^{c}v_{i}}{N}+\sum_{a}\hat{\lambda}_{a}^{\mu}\frac{\sum_{i=1}^{\eta N}w_{i}^{a}v_{i}}{N}\right)}\times \\ &  e^{-\frac{\Delta}{2}\left(\sum_{cd}\hat{\tilde{\lambda}}_{c}^{\mu}\hat{\tilde{\lambda}}_{d}^{\mu}\frac{\sum_{i=1}^{\eta N}\tilde{w}_{i}^{c}\tilde{w}_{i}^{d}}{N}+\sum_{ab}\hat{\lambda}_{a}^{\mu}\hat{\lambda}_{b}^{\mu}\frac{\sum_{i=1}^{\eta N}w_{i}^{a}w_{i}^{b}}{N}+2\sum_{ac}\hat{\lambda}_{a}^{\mu}\hat{\tilde{\lambda}}_{c}^{\mu}\frac{\sum_{i=1}^{\eta N}w_{i}^{a}\tilde{w}_{i}^{c}}{N}\right)},
\end{eqnarray}
and the zeros components:
\begin{eqnarray}
 &  & \prod_{i=\eta N+1}^{N}\mathbb{E}_{x_{i}^{\mu}}e^{-i\left(\sum_{c}\hat{\tilde{u}}_{c}^{\mu}\frac{\tilde{w}_{i}^{c}}{\sqrt{N}}\right)x_{i}} =\\
 & = & e^{-i\left(2y^{\mu}-1\right)\left(\sum_{c}\hat{\tilde{u}}_{c}^{\mu}\frac{\sum_{i=\eta N+1}^{N}\tilde{w}_{i}^{c}v_{i}}{N}\right)-\frac{\Delta}{2}\sum_{cd}\hat{\tilde{u}}_{c}^{\mu}\hat{\tilde{u}}_{d}^{\mu}\frac{\sum_{i=\eta N+1}^{N}\tilde{w}_{i}^{c}\tilde{w}_{i}^{d}}{N}}.
\end{eqnarray}

Finally, we introduce the overlap parameters:
\begin{itemize}
\item $\tilde{m}^{c}=\frac{\sum_{i=1}^{\eta N}\tilde{w}_{i}^{c}v_{i}}{N}$ and
$\tilde{m}_{0}^{c}=\frac{\sum_{i=\eta N+1}^{N}\tilde{w}_{i}^{c}v_{i}}{N}$, representing the magnetization of the teacher in the direction of the true signal $v$.
\item $m^{a}=\frac{\sum_{i=1}^{\eta N}w_{i}^{a}v_{i}}{N}$, representing the magnetization of the student in the direction of $v$.
\item $\tilde{q}^{cd}=\frac{\sum_{i=1}^{\eta N}\tilde{w}_{i}^{c}\tilde{w}_{i}^{d}}{N}$ and  $\tilde{q}_{0}^{cd}=\frac{\sum_{i=\eta N+1}^{N}\tilde{w}_{i}^{c}\tilde{w}_{i}^{d}}{N}$, representing the typical overlap between different teacher configurations. 
\item $q^{ab}=\frac{\sum_{i=1}^{N}w_{i}^{a}w_{i}^{b}}{N}=\frac{\sum_{i=1}^{\eta N}w_{i}^{a}w_{i}^{b}}{N}$, representing the typical overlap between different student configurations. 
\item $S^{ac}=\frac{\sum_{i=1}^{\eta N}w_{i}^{a}\tilde{w}_{i}^{c}}{N}$, representing the typical overlap between teacher and student.
\end{itemize}

We can now rewrite the replicated volume as:
\[
\Omega^{n}=\lim_{\tilde{n}\to0}\int\prod_{c}d\tilde{b}^{c}\int\prod_{a}db^{a}\int\prod_{c}\frac{d\tilde{m}^{c}d\hat{\tilde{m}}^{c}}{2\pi/N}\int\prod_{c}\frac{d\tilde{m}_{0}^{c}d\hat{\tilde{m}}_{0}^{c}}{2\pi/N}\int\prod_{a}\frac{dm^{a}d\hat{m}^{a}}{2\pi/N}\int\prod_{cd}\frac{d\tilde{q}^{cd}d\hat{\tilde{q}}^{cd}}{2\pi/N}\times
\]
\begin{equation}
\times\int\prod_{cd}\frac{d\tilde{q}_{0}^{cd}d\hat{\tilde{q}}_{0}^{cd}}{2\pi/N}\int\prod_{ab}\frac{dq^{ab}d\hat{q}^{ab}}{2\pi/N}\int\prod_{ac}\frac{dS^{ac}d\hat{S}^{ac}}{2\pi/N}G_{I}\left(G_{S}\right)^{\eta N}\left(G_{S_0}\right)^{(1-\eta)N}\left(G_{E}\right)^{\alpha N} \label{eq:stud-action}
\end{equation}
with the definitions for the interaction term:
\begin{equation}
\medmath{G_{I}=\exp\left(-N\left(\sum_{c}\left(\hat{\tilde{m}}^{c}\tilde{m}^{c}+\hat{\tilde{m}}_{0}^{c}\tilde{m}_{0}^{c}\right)+\sum_{a}\hat{m}^{a}m^{a}+\sum_{cd}\left(\hat{\tilde{q}}^{cd}\tilde{q}^{cd}+\hat{\tilde{q}}_{0}^{cd}\tilde{q}_{0}^{cd}\right)+\sum_{ab}\hat{q}^{ab}q^{ab}+\sum_{ca}\hat{S}^{ca}S^{ca}\right)\right)}
\end{equation}
the two entropic terms:
\[ 
G_{S}=\int\prod_{c}d \tilde{w}^{c} e^{-\frac{1}{2}\tilde{\beta}\tilde{\lambda} (\tilde{w}^c)^2}\int\prod_{a}d w^{a} e^{-\frac{1}{2}\beta\lambda (w^a)^2} \times
\]
\begin{equation}
\times\exp\left(\sum_{c}\hat{\tilde{m}}^{c}\tilde{w}^{c}+\sum_{a}\hat{m}^{a}w^{a}+\sum_{cd}\hat{\tilde{q}}^{cd}\tilde{w}^{c}\tilde{w}^{d}+\sum_{ab}\hat{q}^{ab}w^{a}w^{b}+\sum_{ca}\hat{S}^{ca}w^{a}\tilde{w}^{c}\right),
\end{equation}
\begin{equation}
G_{S_0}=\int\prod_{c}d \tilde{w}^{c} e^{-\frac{1}{2}\tilde{\beta}\tilde{\lambda} (\tilde{w}^c)^2}\exp\left(\sum_{c}\hat{\tilde{m}}_{0}^{c}\tilde{w}^{c}+\sum_{cd}\hat{\tilde{q}}_{0}^{cd}\tilde{w}^{c}\tilde{w}^{d}\right),
\end{equation}
and the energetic term:
\[
G_{E}=\mathbb{E}_y\int\prod_{c}\frac{d\tilde{u}^{c}d\hat{\tilde{u}}^{c}}{2\pi}e^{i\hat{\tilde{u}}^{c}\tilde{u}^{c}}\int\prod_{c}\frac{d\tilde{\lambda}^{c}d\hat{\tilde{\lambda}}^{c}}{2\pi}e^{i\hat{\tilde{\lambda}}^{c}\tilde{\lambda}^{c}}\int\prod_{a}\frac{d\lambda^{a}d\hat{\lambda}^{a}}{2\pi}e^{i\lambda^{a}\hat{\lambda}^{a}}\times
\]
\[
\times e^{-\frac{\Delta}{2}\left(\sum_{cd}\hat{\tilde{u}}_{c}\hat{\tilde{u}}_{d}\tilde{q}_{0}^{cd}+\sum_{cd}\hat{\tilde{\lambda}}_{c}\hat{\tilde{\lambda}}_{d}\tilde{q}^{cd}+\sum_{ab}\hat{\lambda}_{a}\hat{\lambda}_{b}q^{ab}+2\sum_{ac}\hat{\lambda}_{a}\hat{\tilde{\lambda}}_{c}S^{ac}\right)}\times
\]
\begin{equation}
\times \prod_{c}e^{-\frac{\tilde{\beta}}{2}\ell\left(y,\tilde{u}^{c}+\tilde{\lambda}^{c}+(2y-1)\left(\tilde{m}_{0}^{c}+\tilde{m}^{c}\right)+\tilde{b}^{c}\right)}\prod_{a}e^{-\frac{\beta}{2}\ell\left(\tilde{u}^{1}+\tilde{\lambda}^{1}+(2y-1)\left(\tilde{m}_{0}^{1}+\tilde{m}^{1}\right)+\tilde{b}^{1},\lambda^{a}+(2y-1)m^{a}+b^{a}\right)}
\end{equation}

\subsubsection{Replica symmetric Ansatz}
Since also the student learning problem entails a convex optimization, we can safely assume Replica Symmetry to be realized also at the level of its order parameters. Moreover, as in the previous calculation, we should average over the realizations of the Gaussian signal $\boldsymbol{v}$ but we can also exploit the isotropy and fix the gauge $\boldsymbol{v}=\boldsymbol{1}^T$. The two magnetizations characterizing the overlap between the teacher vector and the signal, $\tilde{m}$ and $\tilde{m}_0$ (along the first $\eta N$ components and the complementary $(1-\eta)N$ components), will typically give $\tilde{m} / \tilde{m}_0 = \eta / (1-\eta)$. Thus we can pose:
\begin{itemize}
\item For the teacher magnetizations $\tilde{m}^{c}=\eta\,\tilde{m}$, $\tilde{m}_{0}^{c}=(1-\eta)\,\tilde{m}$. We can also set the conjugates to be equal: $\hat{\tilde{m}}^{c}=\hat{\tilde{m}}_{0}^{c}=\hat{\tilde{m}}$. 
\item Similarly, $\tilde{q}^{cd}=\eta\tilde{\,q}$, $\tilde{q}_{0}^{cd}=(1-\eta)\,\tilde{q}$
for $c\neq d$; and $\tilde{q}^{cd}=\eta\,\tilde{Q}$,  $\tilde{q}_{0}^{cd}=\left(1-\eta\right)\,\tilde{Q}$
for $c=d$. 
\item For the student magnetization and self overlap $m^{a}=m$, $q^{ab}=q$ for $a\neq b$; $q^{ab}=Q$ for $a=b$.
\item Since the student is coupled only to the first replica of the teacher, in general we will have two distinct teacher-strudent overlaps: $S^{ca}=S$ for $c=1$, $S^{ca}=\tilde{S}$ for $c\neq1$.
\end{itemize}

It is easy to see that the order parameters referred to the teacher are determined by the same saddle point-equations obtained in the previous section, since at finite $\tilde{n}$ we can send $n\to 0$ and the $\mathcal{O}(1)$ term corresponds to the action Eq.\,(\ref{eq:T-FE}).
In order to get the fixed point equations that charachterize the student, instead, we can substitute the $\tilde{n}\to0$ limit explicitly and keep the $\mathcal{O}(n)$ terms in the series expansion at small $n$ of the action in (\ref{eq:stud-action}). 

In this limit, the normalized logarithm of the interaction term yields:
\begin{eqnarray}
g_{I} = -\left(\hat{m}m+\frac{\hat{Q}Q}{2}-\frac{1}{2}\hat{q}q+\hat{S}S-\hat{\tilde{S}}\tilde{S}\right).
\end{eqnarray}

In the $\tilde{\eta}\to0$ limit, the entropic term $G_{S_0}$ does not contribute to the saddle-point, so it will be ignored in the following. The calculations for $G_S$, instead, are a bit more involved than in the previous case because of the double average characterizing the distillation framework. After some manipulation and
three separate Hubbard-Stratonovich, introducing the Gaussian variables $x,z$ and $\tilde{z}$, one gets:

\begin{eqnarray}
g_{S} & = & \frac{1}{n}  \lim_{\tilde{n}\to0}\log\int\mathcal{D}x\int\mathcal{D}z\int\mathcal{D}\tilde{z}\times\\
 &  & \times\int\prod_{c}d\tilde{w}^{c}\exp\left(\frac{1}{2}\left(\hat{\tilde{Q}}-\hat{\tilde{q}} - \tilde{\lambda}\right)\sum_{c}\left(\tilde{w}^{c}\right)^{2}+\left(\hat{\tilde{m}}+\sqrt{\hat{\tilde{S}}}x+\sqrt{\hat{\tilde{q}}-\hat{\tilde{S}}}\tilde{z}\right)\sum_{c}\tilde{w}^{c}\right)\\
 &  & \medmath{\times\int\prod_{a}d w^{a} \exp\left(\frac{1}{2}\left(\hat{Q}-\hat{q} -\lambda \right)\sum_{a}\left(w^{a}\right)^{2}+\left(\hat{m}+\sqrt{\hat{\tilde{S}}}x+\left(\hat{S}-\hat{\tilde{S}}\right)w_{1}+\sqrt{\hat{q}-\hat{\tilde{S}}}z\right)\sum_{a}w^{a}\right)}\\
 & = & \int\mathcal{D}x\int\mathcal{D}z\int\mathcal{D}\tilde{z}\frac{\int d\tilde{w}\exp\left(\tilde{A}\right)\,\log\left(\int dw\exp\left(A\right)\right)}{\int d\tilde{w}\exp\left(\tilde{A}\right)}
\end{eqnarray}
\normalsize
where by sending $\tilde{n}\to0$ we reestablished the expectation appearing in Eq.\,(\ref{eq:dist_FE}) and where we defined:
\begin{equation}
\tilde{A}=\frac{1}{2}\left(\hat{\tilde{Q}}-\hat{\tilde{q}} - \tilde{\lambda}\right)\left(\tilde{w}\right)^{2}+\left(\hat{\tilde{m}}+\sqrt{\hat{\tilde{S}}}x+\sqrt{\hat{\tilde{q}}-\hat{\tilde{S}}}\tilde{z}\right)\tilde{w}
\end{equation}
\begin{equation}
A=\frac{1}{2}\left(\hat{Q}-\hat{q} -\lambda\right)\left(w\right)^{2}+\left(\hat{m}+\sqrt{\hat{\tilde{S}}}x+\left(\hat{S}-\hat{\tilde{S}}\right)w_{1}+\sqrt{\hat{q}-\hat{\tilde{S}}}z\right)w.
\end{equation}

After a couple rotations between the introduced Gaussian variables, it is possible to perform the $\int\mathcal{D}x$ integral analytically and obtain:
\begin{equation}
g_{s}/n=\int\mathcal{D}z\int\mathcal{D}\tilde{z}\frac{\int d\tilde{w}\exp\left(\tilde{A}\right)\,\log\left(\int dw\exp\left(A\right)\right)}{\int d\tilde{w}\exp\left(\tilde{A}\right)}
\end{equation}
with:
\begin{equation}
\tilde{A}^{\prime}=\frac{1}{2}\left(\hat{\tilde{Q}}-\hat{\tilde{q}} -\tilde{\lambda}\right)\left(\tilde{w}\right)^{2}+\left(\hat{\tilde{m}}+\sqrt{\hat{\tilde{q}}}\tilde{z}\right)\tilde{w}
\end{equation}
\begin{equation}
A^{\prime}=\frac{1}{2}\left(\hat{Q}-\hat{q} -\lambda\right)\left(w\right)^{2}+\left(\hat{m}+\left(\hat{S}-\hat{\tilde{S}}\right)w_{1}+\frac{\hat{\tilde{S}}}{\sqrt{\hat{\tilde{q}}}}\tilde{z}+\sqrt{\hat{q}-\frac{\hat{\tilde{S}}^{2}}{\hat{\tilde{q}}}}z\right)w.
\end{equation}

The calculation follows the same lines also for the energetic term, where after the introduction of four $x,\tilde{z},\hat{z}$ and $z$, one can factorize over the replica indices and send $\tilde{n}\to0$, yielding: 

\begin{equation}
    g_{E} =  \mathbb{E}_y \int\mathcal{D}x\int\mathcal{D}\tilde{z}\int\mathcal{D}\hat{z}\int\mathcal{D}z \times \nonumber
\end{equation}
\begin{equation}
\medmath{\frac{\int\frac{d\tilde{u}d\hat{\tilde{u}}}{2\pi}\int\frac{d\tilde{\lambda}d\hat{\tilde{\lambda}}}{2\pi}e^{\tilde{B}_{0}+\tilde{B}-\frac{\tilde{\beta}}{2}\ell\left(y,\sigma\left(\tilde{u}+\tilde{\lambda}+(2y-1)\tilde{m}+\tilde{b}\right)\right)}\log\int\frac{d\lambda d\hat{\lambda}}{2\pi}e^{B-\frac{\beta}{2}\ell\left(\sigma\left(\tilde{u}+\tilde{\lambda}+(2y-1)\tilde{m}+\tilde{b}\right),\sigma\left(\lambda+(2y-1)m+b\right)\right)}}{\int\frac{d\tilde{u}d\hat{\tilde{u}}}{2\pi}\int\frac{d\tilde{\lambda}d\hat{\tilde{\lambda}}}{2\pi}e^{\tilde{B}_{0}+\tilde{B}-\frac{\tilde{\beta}}{2}\ell\left(y,\sigma\left(\tilde{u}+\tilde{\lambda}+(2y-1)\tilde{m}+\tilde{b}\right)\right)}}} 
\end{equation}
where we defined:
\begin{equation}
\tilde{B}_{0}=-\frac{\Delta}{2}\left(1-\eta\right)\left(\tilde{Q}-\tilde{q}\right)\hat{\tilde{u}}^{2}+i\hat{\tilde{u}}\left(\tilde{u}+\sqrt{\Delta\left(1-\eta\right)\tilde{q}}\hat{z}\right)
\end{equation}
\begin{equation}
\tilde{B}=-\frac{\Delta}{2}\eta\left(\tilde{Q}-\tilde{q}\right)\left(\hat{\tilde{\lambda}}\right)^{2}+i\hat{\tilde{\lambda}}\left(\tilde{\lambda}+\sqrt{\Delta\tilde{S}}x+\sqrt{\Delta\left(\eta\tilde{q}-\tilde{S}\right)}\tilde{z}\right)
\end{equation}
\begin{equation}
B=-\frac{\Delta}{2}\left(Q-q\right)\hat{\lambda}^{2}+i\hat{\lambda}\left(\lambda+\sqrt{\Delta\tilde{S}}x+\sqrt{\Delta\left(q-\tilde{S}\right)}z+i\Delta\left(S-\tilde{S}\right)\hat{\tilde{\lambda}}\right)
\end{equation}

In order to disentangle the Gaussian integrations and allow us to land on an expression where we can explicitly perform a few of them we start by shifting $z^{\prime}=z+i\hat{\tilde{\lambda}}\frac{\Delta\left(S-\tilde{S}\right)}{\sqrt{\Delta\left(q-\tilde{S}\right)}}$. Now, we can proceed and simplify the $d\hat{\lambda},d\hat{\tilde{\lambda}},d\hat{\tilde{u}}$ integrals and get, after shifting and rescaling $d\lambda,d\tilde{\lambda},d\tilde{u}$:
\begin{equation}
g_{E}=\mathbb{E}_y \int\mathcal{D}x\int\mathcal{D}\tilde{z}\int\mathcal{D}\hat{z}\frac{\int\mathcal{D}z\int\mathcal{D}\tilde{u}\int\mathcal{D}\tilde{\lambda}e^{-\frac{\tilde{\beta}}{2}\ell\left(y,\sigma\left(\tilde{h}^{\prime}\right)\right)}\log\int\mathcal{D}\lambda e^{-\frac{\beta}{2}\ell\left(\sigma\left(\tilde{h}^{\prime}\right),\sigma\left(h\right)\right)}}{\int\mathcal{D}\tilde{u}\int\mathcal{D}\tilde{\lambda}e^{-\frac{\tilde{\beta}}{2}\ell\left(y,\sigma\left(\tilde{h}\right)\right)}},
\end{equation}
with:
\[
\medmath{\tilde{h}^{\prime}=\sqrt{\Delta\left(1-\eta\right)\left(\tilde{Q}-\tilde{q}\right)}\tilde{u}-\sqrt{\Delta\left(1-\eta\right)\tilde{q}}\hat{z}+\sqrt{\Delta\left(\eta\left(\tilde{Q}-\tilde{q}\right)-\frac{\left(S-\tilde{S}\right)^{2}}{q-\tilde{S}}\right)}\tilde{\lambda}+}
\]
\begin{equation}
\medmath{-\left(\sqrt{\Delta\tilde{S}}x+\sqrt{\Delta\left(\eta\tilde{q}-\tilde{S}\right)}\tilde{z}+\frac{\Delta\left(S-\tilde{S}\right)}{\sqrt{\Delta\left(q-\tilde{S}\right)}}z\right)+\left(2y-1\right)\,\tilde{m}+\tilde{b}},
\end{equation}

\begin{equation}
\medmath{\tilde{h}=\sqrt{\Delta\left(1-\eta\right)\left(\tilde{Q}-\tilde{q}\right)}\tilde{u}-\sqrt{\Delta\left(1-\eta\right)\tilde{q}}\hat{z}+\sqrt{\Delta\eta\left(\tilde{Q}-\tilde{q}\right)}\tilde{\lambda}-\left(\sqrt{\Delta\tilde{S}}x+\sqrt{\Delta\left(\eta\tilde{q}-\tilde{S}\right)}\tilde{z}\right)+\left(2y-1\right)\,\tilde{m}+\tilde{b}}
\end{equation}
\begin{equation}
h=\sqrt{\Delta\left(Q-q\right)}\lambda-\left(\sqrt{\Delta\tilde{S}}x+\sqrt{\Delta\left(q-\tilde{S}\right)}z\right)+\left(2y-1\right)\,m+b.
\end{equation}

After a few rotations between $\tilde{\lambda},\tilde{u},z,\hat{z},\tilde{z}$ and $x$, one can perform the $\mathcal{D}\tilde{u}, \mathcal{D}\hat{z}, \mathcal{D}x$ integrals and get the final expression:
\begin{equation}
g_{E}/n=\left\langle \int\mathcal{D}z\int\mathcal{D}\tilde{z}\frac{\int\mathcal{D}\tilde{\lambda}e^{-\frac{\tilde{\beta}}{2}\ell\left(y,\sigma\left(\tilde{h}\right)\right)}\log\int\mathcal{D}\lambda e^{-\frac{\beta}{2}\ell\left(\sigma\left(\tilde{h}\right),\sigma\left(h\right)\right)}}{\int\mathcal{D}\tilde{\lambda}e^{-\frac{\tilde{\beta}}{2}\ell\left(y,\sigma\left(\tilde{h}\right)\right)}}\right\rangle _{y}
\end{equation}
with the definitions:
\begin{equation}
\tilde{h}=\sqrt{\Delta\left(\tilde{Q}-\tilde{q}\right)}\tilde{\lambda}+\sqrt{\Delta\tilde{q}}\tilde{z}+\left(2y-1\right)\,\tilde{m}+\tilde{b},
\end{equation}
\begin{equation}
h=\sqrt{\Delta\left(Q-q\right)}\lambda+\frac{\Delta\left(S-\tilde{S}\right)\tilde{\lambda}}{\sqrt{\Delta\left(\tilde{Q}-\tilde{q}\right)}}+\frac{\sqrt{\Delta}\tilde{S}\tilde{z}}{\sqrt{\tilde{q}}}+\sqrt{\Delta\left(q-\frac{\tilde{S}^{2}}{\tilde{q}}-\frac{\left(S-\tilde{S}\right)^{2}}{\left(\tilde{Q}-\tilde{q}\right)}\right)}z+\left(2y-1\right)\,m+b.
\end{equation}

\subsubsection{Zero temperature limit}
Finally, we can recover the nested optimization characterizing the distillation framework by successively taking the two limits $\tilde{\beta}\to\infty$ and $\beta\to\infty$. As in the previous calculation, we have to introduced rescaled overlap order parameters:
\begin{equation}
\left(\tilde{Q}-\tilde{q}\right)=\delta \tilde{q}/\tilde{\beta}, \,\,\,\,\,\, \left(Q-q\right)=\delta q/\beta, \,\,\,\,\,\, S-\tilde{S}=\delta S/\tilde{\beta}
\end{equation}
and rescale also all the conjugate parameters:
\begin{itemize}
\item $\hat{\tilde{Q}}\to\tilde{\beta}^{2}\hat{\tilde{q}}+\mathcal{O}\left(\tilde{\beta}\right)$,
$\hat{\tilde{q}}\to\tilde{\beta}^{2}\hat{\tilde{q}}$, $(\hat{\tilde{Q}}-\hat{\tilde{q}})\to-\tilde{\beta}\delta\hat{\tilde{q}}$, $\hat{\tilde{m}}\to\tilde{\beta}\hat{\tilde{m}}$,
\item $\hat{Q}\to\beta^{2}\hat{q}+\mathcal{O}\left(\beta\right)$, $\hat{q}\to\beta^{2}\hat{q}$, $(\hat{Q}-\hat{q})\to-\beta\delta\hat{q}$, $\hat{m}\to\beta\hat{m}$,
\item $\hat{S}\to\tilde{\beta}\beta\hat{S}+\mathcal{O}\left(\beta\right)$, $\hat{\tilde{S}}\to\tilde{\beta}\beta\hat{S}$, $(\hat{S}-\hat{\tilde{S}})\to\beta\delta\hat{S}$.
\end{itemize}
With these scalings, the interaction term reads:
\begin{equation}
g_{I} = -\beta\left(\hat{m}m+\frac{1}{2}\left(\hat{q}\delta q-\delta\hat{q}q\right)+\left(\hat{S}\delta S+\delta\hat{S}\tilde{S}\right)\right)+\mathcal{O}\left(1\right).
\end{equation}
In the entropic term, in the zero-temperature limit the integrals over the teacher and student weights become extremum operations:
\begin{equation}
g_{S}=\lim_{\beta\to\infty}\beta\int\mathcal{D}z\int\mathcal{D}\tilde{z}\,\,M_{s}^{\star}
\end{equation}
where:
\[
M_{s}^{\star}=\mathrm{max}_{w}\left\{ -\frac{1}{2}\left(\lambda+\delta\hat{q}\right)w^{2}+\left(\hat{m}+\delta\hat{S}\tilde{w}^{\star}+\frac{\hat{S}}{\sqrt{\hat{\tilde{q}}}}\tilde{z}+\sqrt{\frac{\hat{\tilde{q}}\hat{q}-\hat{S}^{2}}{\hat{\tilde{q}}}}z\right)w\right\} 
\]
\begin{equation}
=\frac{1}{2}\frac{\left(\hat{m}+\delta\hat{S}\tilde{w}^{\star}+\frac{\hat{S}}{\sqrt{\hat{\tilde{q}}}}\tilde{z}+\sqrt{\frac{\hat{\tilde{q}}\hat{q}-\hat{S}^{2}}{\hat{\tilde{q}}}}z\right)^{2}}{\lambda+\delta\hat{q}}
\end{equation}
and where the teacher weight configuration, as in Eq.~(\ref{eq:teach_max}), maximizes the action:
\begin{equation} 
\tilde{w}^{\star}=\mathrm{argmax}_{\tilde{w}}\left\{ -\frac{1}{2}(\tilde{\lambda}+\,\delta\hat{\tilde{q}})\tilde{w}^{2}+(\hat{\tilde{m}}+\sqrt{\hat{\tilde{q}}}\tilde{z})\tilde{w}\right\} =\frac{\hat{\tilde{m}}+\sqrt{\hat{\tilde{q}}}\tilde{z}}{\tilde{\lambda}+\delta\hat{\tilde{q}}}.
\end{equation}
With an analytic expression for the maxima, the $\int\mathcal{D}z\int\mathcal{D}\tilde{z}$ integrations
can be carried out, giving:
\begin{equation}
g_S=\frac{\beta}{2}\frac{\left(\hat{m}+\hat{\tilde{m}}\frac{\delta\hat{S}}{\tilde{\lambda}+\delta\hat{\tilde{q}}}\right)^{2}+\left(\frac{\hat{S}}{\sqrt{\hat{\tilde{q}}}}+\frac{\delta\hat{S}\sqrt{\hat{\tilde{q}}}}{\tilde{\lambda}+\delta\hat{\tilde{q}}}\right)^{2}+\frac{\hat{\tilde{q}}\hat{q}-\hat{S}^{2}}{\hat{\tilde{q}}}}{\lambda+\delta\hat{q}}.
\end{equation}

Lastly, in the $\tilde{\beta},\beta\to\infty$, also the $\mathcal{D}\lambda, \mathcal{D}\tilde{\lambda}$ integrals in the energetic term become one-dimensional extremum operations, and we have:
\begin{equation}
g_{E}=\beta\mathbb{E}_y \int\mathcal{D}z\int\mathcal{D}\tilde{z}\,M_{E}^{\star}
\end{equation}
where:
\begin{equation}
M_{E}^{\star}=\max_{\lambda}\left\{ -\frac{1}{2}\frac{\lambda^{2}}{\Delta\,\delta q}-\frac{1}{2}\ell\left(\sigma\left(\tilde{h}\left(\lambda^{\star}\right)\right),\sigma\left(h\left(\lambda\right)\right)\right)\right\} 
\end{equation}
with:
\begin{equation}
\tilde{h}\left(\tilde{\lambda}\right)=\tilde{\lambda}+\left(2y-1\right)\,\tilde{m}+\tilde{b}+\sqrt{\Delta\tilde{q}}\tilde{z}
\end{equation}
\begin{equation}
h\left(\lambda\right)=\lambda+\left(2y-1\right)\,m+b+\frac{\delta S}{\delta\tilde{q}}\tilde{\lambda}+\sqrt{\Delta}\frac{S}{\sqrt{\tilde{q}}}\tilde{z}+\sqrt{\Delta\left(q-\frac{S^{2}}{\tilde{q}}+\mathcal{O}\left(\frac{1}{\tilde{\beta}}\right)\right)}z
\end{equation}
and
\begin{equation}
\tilde{\lambda}^{\star}=\mathrm{argmax}_{\tilde{\lambda}}-\frac{1}{2}\frac{\tilde{\lambda}^{2}}{\Delta\,\delta\tilde{q}}-\frac{1}{2}\ell\left(y,\sigma\left(\tilde{h}\left(\tilde{\lambda}\right)\right)\right).
\end{equation}

The free-entropy of Eq.\,(\ref{eq:dist_finalFE}) is thus recovered after dividing the various contributions by $\beta$:
\begin{equation}
\Phi=-\left(\hat{m}m+\frac{1}{2}\left(\hat{q}\delta q-\delta\hat{q}q\right)+\left(\hat{S}\delta S+\delta\hat{S}S\right)\right)+\eta\,g_{S}+\alpha\,g_{E}
\end{equation}

\subsection{Distillation with optimal teacher}

The replica calculation for the distillation setting with optimal teacher is very similar to the one presented in the first section of the Appendix (for the typical logistic regression framework), since we make an explicit assumption on the statistical measure for the teacher weight vector and the double average (as in the previous section) is not needed. 

As described in the main text, we assume the teacher to be represented by a noisy version of the signal $\boldsymbol{w} = \boldsymbol{v} + \sqrt{\frac{\Delta}{\alpha}} \boldsymbol{h}$, where each component of the noise is independent and normal distributed $h_i\sim\mathcal{N}(0,1)$. This choice induces an average magnetization $\tilde{m}=1+\mathcal{O}(N^{-1/2})$ and a norm $\tilde{q}=1+\frac{\Delta}{\alpha}$, same as in the case of the optimal plug-in estimator of Eq.\,(\ref{eq:plug-in}). When the bias is set to $\tilde{b}=\Delta\frac{(1+\Delta/\alpha)}{2} \log(\frac{\rho}{1-\rho})$ the teacher achieves a generalization performance matching the Bayes optimal generalization, and this justifies our modeling choice for characterizing distillation from an optimal teacher.

We thus want to evaluate the free-entropy for an $\eta$-sparse student learning from the outputs produced by this optimal teacher:
\begin{equation}
\medmath{\Phi=\lim_{\beta\to\infty}\frac{1}{\beta N}\left\langle \log\int d\boldsymbol{w}_\eta e^{-\frac{\lambda}{2} \lVert\boldsymbol{w}\rVert_\eta^2}\int db\prod_{\mu}e^{-\beta\,\ell\left(\sigma\left(\frac{\left(\boldsymbol{v}+\boldsymbol{h}\sqrt{\Delta/\alpha}\right)\cdot \boldsymbol{x}^{\mu}}{\sqrt{N}}+\frac{\Delta\left(1+\Delta/\alpha\right)}{2}\log\frac{\rho}{1-\rho}\right),\sigma\left(\frac{\boldsymbol{w}\cdot\boldsymbol{x}^{\mu}}{\sqrt{N}}+b\right)\right)}\right\rangle _{\left\{ \boldsymbol{x}^{\mu},y^{\mu},\boldsymbol{v},\boldsymbol{h}\right\} }}.
\end{equation}
As usual, instead of actually averaging over $\boldsymbol{v}$ we will set $\boldsymbol{v}=\boldsymbol{1}^T$. We can isolate the dependency over the training set by introducing the preactivation variables $l^\mu$ (for the teacher) and $\lambda^\mu_a$ (for the student) via Dirac's delta functions and then perform the disorder average:
\begin{eqnarray}
& &\mathbb{E}_{\boldsymbol{x}^{\mu}}e^{-i\sum_{a}\hat{\lambda}_{a}^{\mu}\frac{\boldsymbol{x}^{\mu}\cdot\boldsymbol{w}^{a}}{\sqrt{N}}-i\hat{l}_{\mu}\frac{\left(\boldsymbol{v}_{i}+\boldsymbol{h}_{i}\sqrt{\Delta/\alpha}\right)\cdot\boldsymbol{x}^{\mu}}{\sqrt{N}}}= \\
& &=e^{-i\left(2y^{\mu}-1\right)\left(\sum_{a}\hat{\lambda}_{a}^{\mu}\frac{\sum_{i=1}^{\eta N}w_{i}^{a}}{N}+\hat{l}_{\mu}\left(1+\sqrt{\Delta/\alpha}\frac{\sum_{i=1}^{N}h_{i}}{N}\right)\right)} \times \\ \nonumber
& &\,\,\,\times e^{-\frac{\Delta}{2}\left(\sum_{ab}\hat{\lambda}_{a}^{\mu}\hat{\lambda}_{b}^{\mu}\frac{\sum_{i=1}^{\eta N}w_{i}^{a}w_{i}^{b}}{N}+2\hat{l}_{\mu}\sum_{a}\hat{\lambda}_{a}^{\mu}\left(\frac{\sum_{i=1}^{\eta N}w_{i}^{a}}{N}+\sqrt{\Delta/\alpha}\frac{\sum_{i=1}^{\eta N}w_{i}^{a}h_{i}}{N}\right)+\hat{l}_{\mu}^{2}\left(1+\Delta/\alpha\frac{\lVert h\rVert^{2}}{N}+2\sqrt{\Delta/\alpha}\frac{\sum_{i=1}^{N}h_{i}}{N}\right)\right)},
\end{eqnarray}
which leads us to introducing the overlap order parameters:
\begin{equation}
m^{a}=\frac{\sum w_{i}^{a}}{N}, \,\,\,\, S^{a}=\frac{\sum w_{i}^{a}h_{i}}{N}, \,\,\,\, q^{ab}=\frac{\sum_{i}w_{i}^{a}w_{i}^{b}}{N}.
\end{equation}
On the other hand, we know that for a Gaussian i.i.d. normal vector $\boldsymbol{h}$ the magnetization in the direction of the signal and the norm will be $\tilde{m}=\frac{\sum h_{i}}{N}=0$ and $\tilde{q}=\frac{\sum h_{i}^{2}}{N}=1$.

The replicated volume thus reads:
\begin{equation}
\Omega^{n}=\int\prod_{a}\frac{d\tilde{S}^{a}d\hat{\tilde{S}}^{a}}{2\pi/N}\int\prod_{a}\frac{dm^{a}d\hat{m}^{a}}{2\pi/N}\int\prod_{ab}\frac{dq^{ab}d\hat{q}^{ab}}{2\pi/N}\int\prod_a db^{a}G_{I}\left(G_{S}\right)^{\eta N}\left(G_{E}\right)^{\alpha N}
\end{equation}
with the definitions:
\begin{equation}
G_{I}=\exp\left(-N\left(\sum_{a}\hat{m}^{a}m^{a}+\sum_{a}\hat{\tilde{S}}^{a}\tilde{S}^{a}+\sum_{ab}\hat{q}^{ab}q^{ab}\right)\right)
\end{equation}
\begin{equation}
G_{S}=\int\mathcal{D}h\int\prod_{a}d\mu\left(w^{a}\right)\exp\left(\sum_{a}\hat{m}^{a}w^{a}+\sum_{a}\hat{S}^{a}w^{a}h+\sum_{ab}\hat{q}^{ab}w^{a}w^{b}\right)
\end{equation}
and (after a little algebra):
\[ 
G_{E}=\int\mathcal{D}l\int\prod_{a}\left(\frac{d\lambda^{a}d\hat{\lambda}^{a}}{2\pi}e^{i\lambda^{a}\hat{\lambda}^{a}}\right)\times
\] 
\[ 
\times e^{-\frac{\Delta}{2}\sum_{ab}\hat{\lambda}_{a}\hat{\lambda}_{b}\left(q^{ab}-\frac{\left(m^{a}+\sqrt{\Delta/\alpha}S^{a}\right)\left(m^{b}+\sqrt{\Delta/\alpha}S^{b}\right)}{1+\Delta/\alpha}\right)-i\,l\sqrt{\frac{\Delta}{\left(1+\Delta/\alpha\right)}}\sum_{a}\hat{\lambda}_{a}\left(m^{a}+\sqrt{\Delta/\alpha}S^{a}\right)}\times 
\]
\begin{equation}
 \times\mathbb{E}_y \prod_{a}e^{-\beta\,\ell\left(\sigma\left(l\sqrt{\Delta\left(1+\Delta/\alpha\right)}+(2y-1)+\frac{\Delta(1-\Delta/\alpha)}{2}\log\frac{\rho}{1-\rho}\right),\sigma\left(\lambda^{a}+(2y-1)m^{a}+b^{a}\right)\right)}
\end{equation}

\subsubsection{RS Ansatz and zero temperature limit}

The rest of the computation is basically identical to that presented in the first section of the Appendix, so we will only report the final expressions for the free-entropy:
\begin{equation}
\Phi=g_I+\eta\,g_{S}+\alpha\,g_{E}.
\end{equation}
with the following definitions for the interaction, entropic and energetic terms:
\begin{equation}
g_I=-\left(\hat{m}m+\hat{S}S+\frac{1}{2}\left(\hat{q}\delta q-\delta\hat{q}q\right)\right),
\end{equation}
\begin{equation}
g_{S}=\frac{\hat{m}^{2}+\hat{S}^{2}+\hat{q}}{2\left(\lambda+\delta\hat{q}\right)},
\end{equation}
\begin{equation}
g_{E}=\mathbb{E}_y \int\mathcal{D}\tilde{z}\int\mathcal{D}z\,\,\, M_{E},
\end{equation}
where:
\begin{equation}
M_{E}=\max_{u}-\frac{u^{2}}{2}-\ell\left(\sigma(\tilde{h}(\tilde{z})),\sigma(h(u,z,\tilde{z}))\right),
\end{equation}
and:
\begin{equation}
\tilde{h}(\tilde{z})=(2y-1)+\frac{\Delta'}{2}\log\frac{\rho}{1-\rho}+\sqrt{\Delta\left(1-\Delta/\alpha\right)}\tilde{z},
\end{equation}
\begin{equation}
\medmath{h(u,z,\tilde{z})=\sqrt{\Delta\delta q}u+\sqrt{\Delta\left(q-\frac{\left(m+\sqrt{\Delta/\alpha}S\right)^{2}}{1+\Delta/\alpha}\right)}z+\sqrt{\frac{\Delta}{\left(1+\Delta/\alpha\right)}}\left(m+\sqrt{\Delta/\alpha}S\right)\tilde{z}+\left(2y-1\right)\,m+b}.
\end{equation}

\section{On the inheritance of the teacher regularization}
\label{sec:more_on_inheritance}

\subsection{Direct and inherited student regularization}

We here show additional experiments on the different effects of direct and indirect (inherited through KD) $L_2$ regularization on the student generalization performance.

\begin{figure}[ht]
\centering
\includegraphics[width=0.5\textwidth]{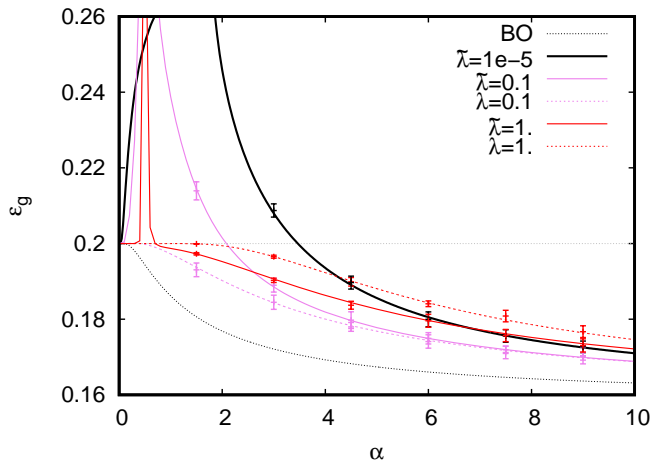}
\caption{\label{fig:Adist_vs_reg} Comparison between the pure distillation generalization curves (blue full curves), as in Fig.\,\ref{fig:dist_vs_reg}, and the performance of a ridge regularized student learning from the labels (red dashed curves), at $\rho=0.2$, $\Delta=1$. (Grey curve) Optimal generalization, achieved with the sparsified plug-in estimator. The data points with error bars represent the results of numerical experiments at $N=4000$ ($10$ samples per point).}
\end{figure}

In Fig.\,\ref{fig:Adist_vs_reg}, we display the distillation generalization curves (full lines) and the direct regularization curves (dashed lines) at fixed values of the regularizer intensity. It is evident that the optimal regularization for the student (with sparsity $\eta=0.5$) is of the same order of the optimal value for the teacher $\lambda\simeq0.1$ (cfr. Fig.\,\ref{fig:typ_learning}). 
If we look at the low $\alpha$ regime, it is also clear that only a direct ridge regularization on the student can grant him a good generalization performance (in the pure distillation setting we see the $\alpha_I$ and $\alpha_S$ interpolation peaks). This observation strongly motivates a mixed approach, where the distillation student is also regularized with an $L_2$ penalty (we will consider this setting in the following).
However, in the large $\alpha$ regime we observe an interesting advantage of distillation: in the case of an overshoot in the regularization intensity ($\lambda\simeq 1$), the distillation student performance is less hindered than the directly regularized student. This provides an indication that the knowledge distillation process may require less fine-tuning than usual empirical risk minimization with cross-entropy loss.

\subsection{On the KD mixing parameter}

We here show some more details on the effect of the mixing parameter $\chi$ in the distillation loss. 

\begin{figure}[ht]
\centering
\includegraphics[width=0.5\textwidth]{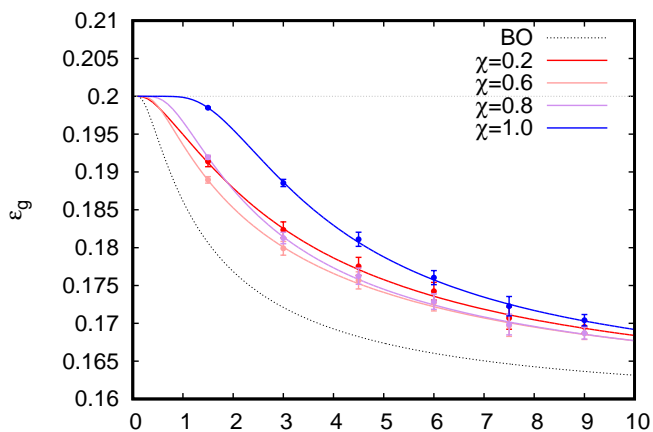}
\caption{\label{fig:Achi} Generalization performance achieved by a  regularized distillation student ($\eta=0.5$, $\lambda=1e-1$, $T=2$) learning from a optimally regularized teacher ($\tilde{\lambda}=0.15$), at varying values of the mixing parameter $\chi$ and with $\rho=0.2$, $\Delta=1$. (Dashed black curve) Generalization bound, achieved by the sparsified plug-in estimator. The data points with error bars represent the results of numerical experiments at $N=4000$ ($10$ samples per point).}
\end{figure}

In Fig.\,\ref{fig:Achi} we fix $\tilde{\lambda},\lambda=0.15$ and vary the value of the mixing parameter $\chi$. As expected from the results reported in the previous paragraphs, we can see that in the low $\alpha$ regime it is better not to rely upon the teacher outputs in order to avoid the interpolation cusp (around $\alpha=\eta$): lower values of $\chi$ (red, pink curves) yield better generalization. As $\alpha$ increases, an appropriate value of the mixing parameter can increase the overall regularization felt by the student up to the optimal amount, guaranteeing an improved performance (purple curve). Interestingly, through the tuning of $\chi$ KD can be made to match the performance achieved with optimal regularization, however a reduction in the performance gap with respect to the plug-in estimator bound is never observed. 

\subsection{Regularization through uniform label smoothing}

We consider a different type of regularization scheme called uniform label smoothing \cite{szegedy2016rethinking, muller2019does}. We introducing a smoothing parameter $\epsilon$ and replacing the "hard" ground truth labels with their softer counterpart $y\to y\left(1-\epsilon\right)+\left(1-y\right)\epsilon$ at training. This type of regularization strategy is known to be effective in preventing overconfidence of the trained model, especially in the case of noisy data. 

\begin{figure}[ht] 
\centering
\includegraphics[width=0.5\textwidth]{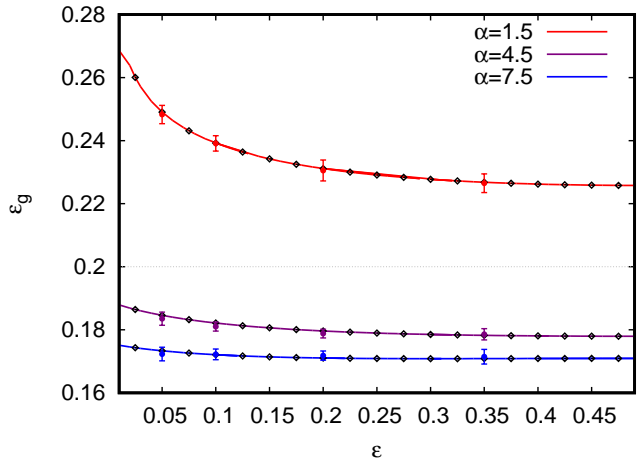}
\caption{ \label{fig:Aregularizations} Comparison of the generalization performance at fixed values of $\alpha$ of a $\eta=0.5$ sparse student in two scenarios: either the student learns directly the smoothened labels (colored lines), or it learns through pure distillation from a teacher trained on the same smoothened labels (black dots). The data points with error bars represent the results of numerical experiments in the second scenario at $N=4000$ ($10$ samples per point).}
\end{figure}

In Fig.\,\ref{fig:Aregularizations}, we compare two different scenarios: in the first case, the student learns from the smoothened labels directly (full colored lines); in the second case, we consider pure distillation from a teacher that learned the smoothened labels (black dots). In both settings the ridge regularization is fixed at the baseline level $\lambda, \tilde{\lambda} = 1e-5$ in order to isolate the regularization effect of the softer labels.  Surprisingly, the student generalization performance obtained in the two scenarios is practically indistinguishable, implying that the KD process can perfectly transfer this type of regularization. Note that, because of the simple nature of the GM generative model, we find it is very beneficial to have learn with higher $\epsilon$ at smaller values of $\alpha$.

\subsection{Varying the Knowledge Distillation temperature}

We consider again a pure distillation setting with $\chi=1$, but now explore the effect of changing the distillation temperature $T$. In particular, in the cross-entropy term in the knowledge distillation loss, the usual outputs will be replaced by $\sigma(h)\to\sigma(h/T)$, where $T$ can increase (lower) the difference between the probabilities of assigning each label.

The general idea behind the introduction of this temperature is that after training, in a multi-class problem, the $softmax$ output function will typically produce small probabilities in correspondence of the incorrect categories and the difference in the assigned weight will be flattened because of the Bolzmann-like form of the activation. Introducing a high temperature can instead reweight the output probability distribution, accentuating the differences in probabilities assigned to incorrect labels \cite{hinton2015distilling, tang2020understanding}. Of course, this effect cannot be explored in the simple binary classification setting. However, it is still possible to observe a positive effect of a high $T$ in the low $\alpha$ regime.
\begin{figure}[ht] 
\centering
\includegraphics[width=0.5\textwidth]{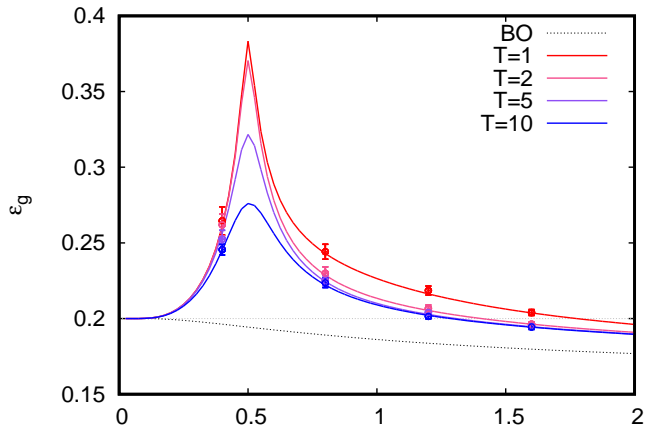}
\caption{ \label{fig:temperatures} Generalization performance in the low $\alpha$ regime of a pure distillation student ($\chi=1$) learning from a optimally regularized teacher ($\tilde{\lambda}=0.15$) with increasing values of the distillation temperature, at $\rho=0.2$, $\Delta=1$, $\eta=0.5$. (Dashed black curve) Generalization bound for the student, achieved by the sparsified plug-in estimator. The data points with error bars represent the results of numerical experiments at $N=4000$ ($10$ samples per point).}
\end{figure}

In Fig.\ref{fig:temperatures} we vary the distillation temperature in the small training set regime, considering an unregularized student that learns from a teacher with ridge regularization intensity $\tilde{\lambda}=0.15$ (nearly optimal setting). It is clear that raising the distillation temperature can mitigate the overfitting phenomenon observed around the $\alpha=\eta$ interpolation peak. Note that, since the magnitude of the learned preactivations is effectively decreased, the saturating regime of the $\sigma(\cdot)$ activation function is avoided and this yields larger differences between the teacher outputs on different patterns, allowing for a better transfer of knowledge.   

\section{On the Double-descent phenomena}
\label{sec:more_on_double-descent}

We have seen in section \ref{sec:double_descent} the appeareance of sharp interpolation peak at $\alpha_I=\eta$, when the number of parameters of the student model equals the size of the training set. Moreover, note that the peak becomes more pronounced when the teacher regularization is close to the optimal value $\tilde{\lambda}=0.1$. 

While such a location for the interpolation peak is uncharacteristic of logistic regression, a similar cusp is typically observed in the weak regularization regime when the classifier is trained with a Mean Squared Error (MSE) loss function \cite{hastie2019surprises,mignacco2020role}. What one would expect in the case of logistic regression is instead a less pronounced peak, located in correspondence of the separability threshold $\alpha_S(\rho, \Delta)$ for the training dataset. Note that, in the mismatched distillation framework, we expect two distinct separability thresholds of this type $\alpha_S<\tilde{\alpha}_S$, one for the student and one for the teacher. 
 
\begin{figure}[ht] 
\centering
\includegraphics[width=0.49\textwidth]{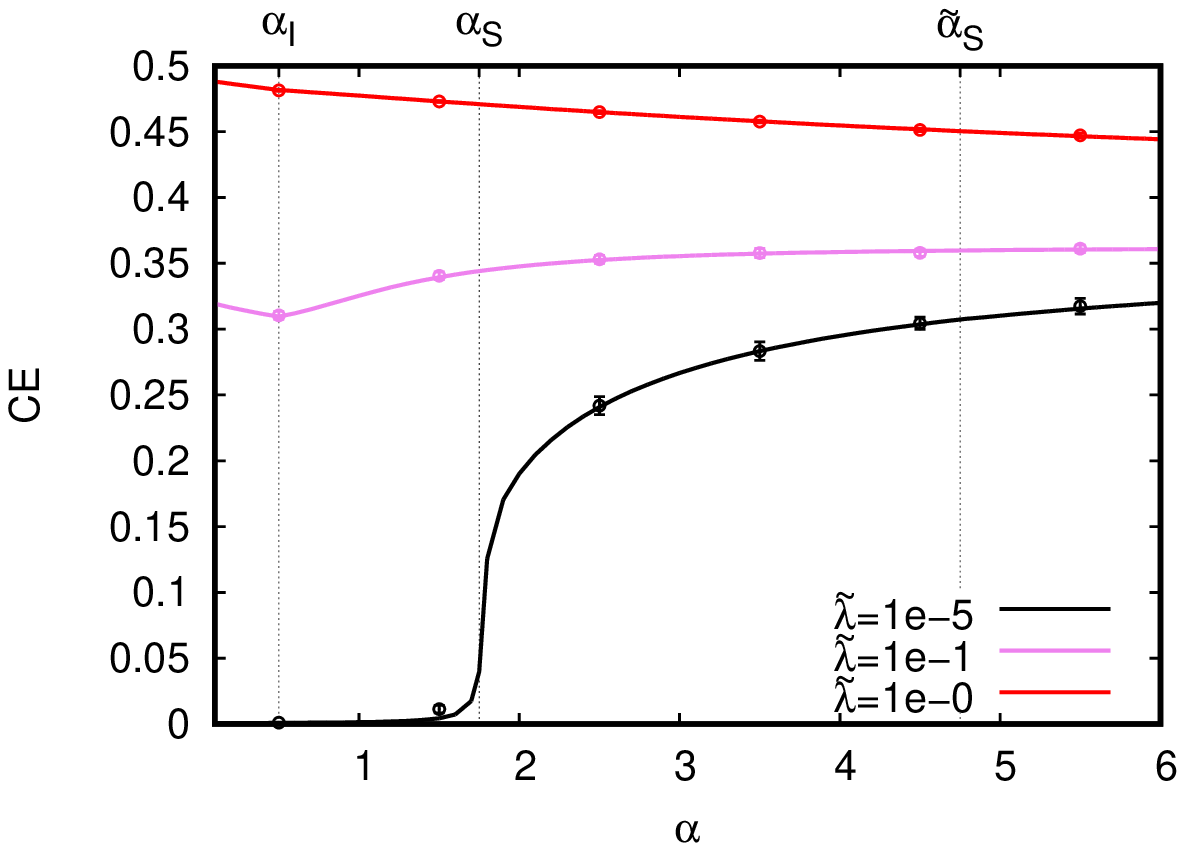}
\includegraphics[width=0.49\textwidth]{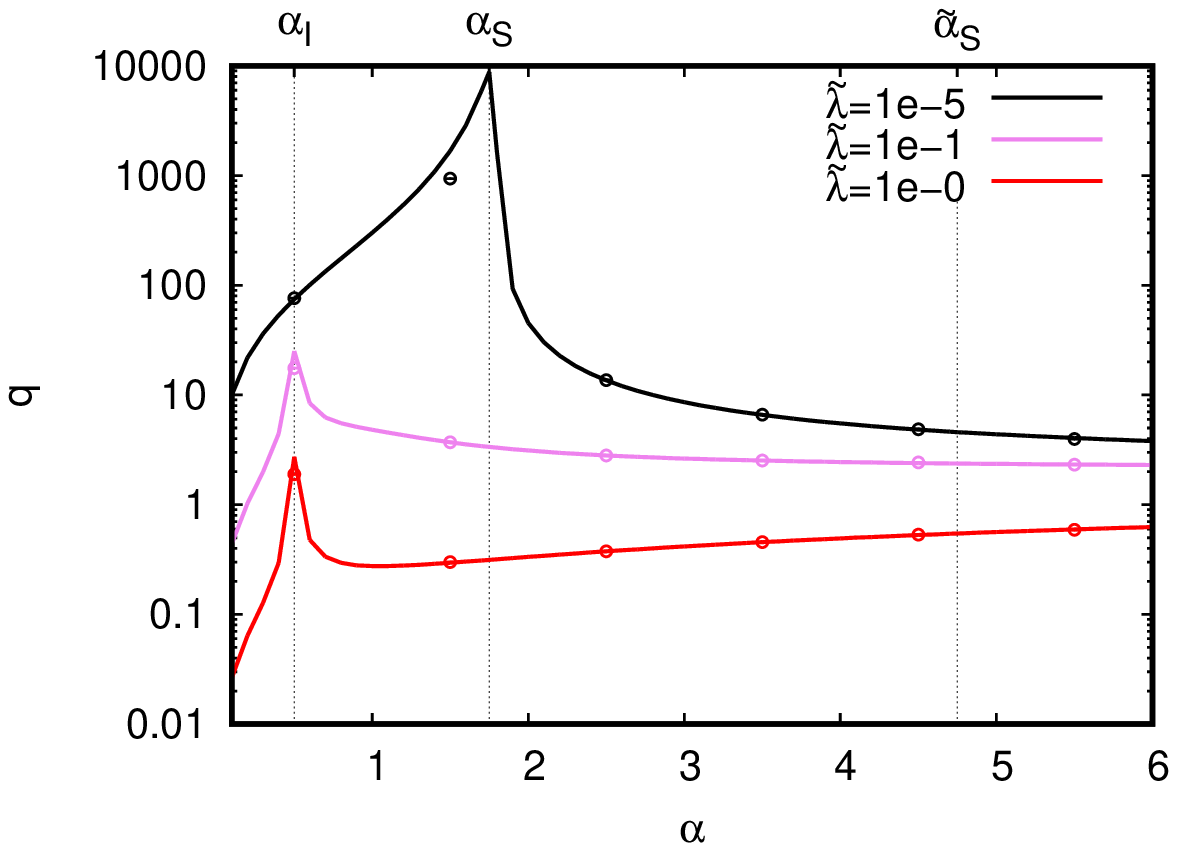}
\includegraphics[width=0.49\textwidth]{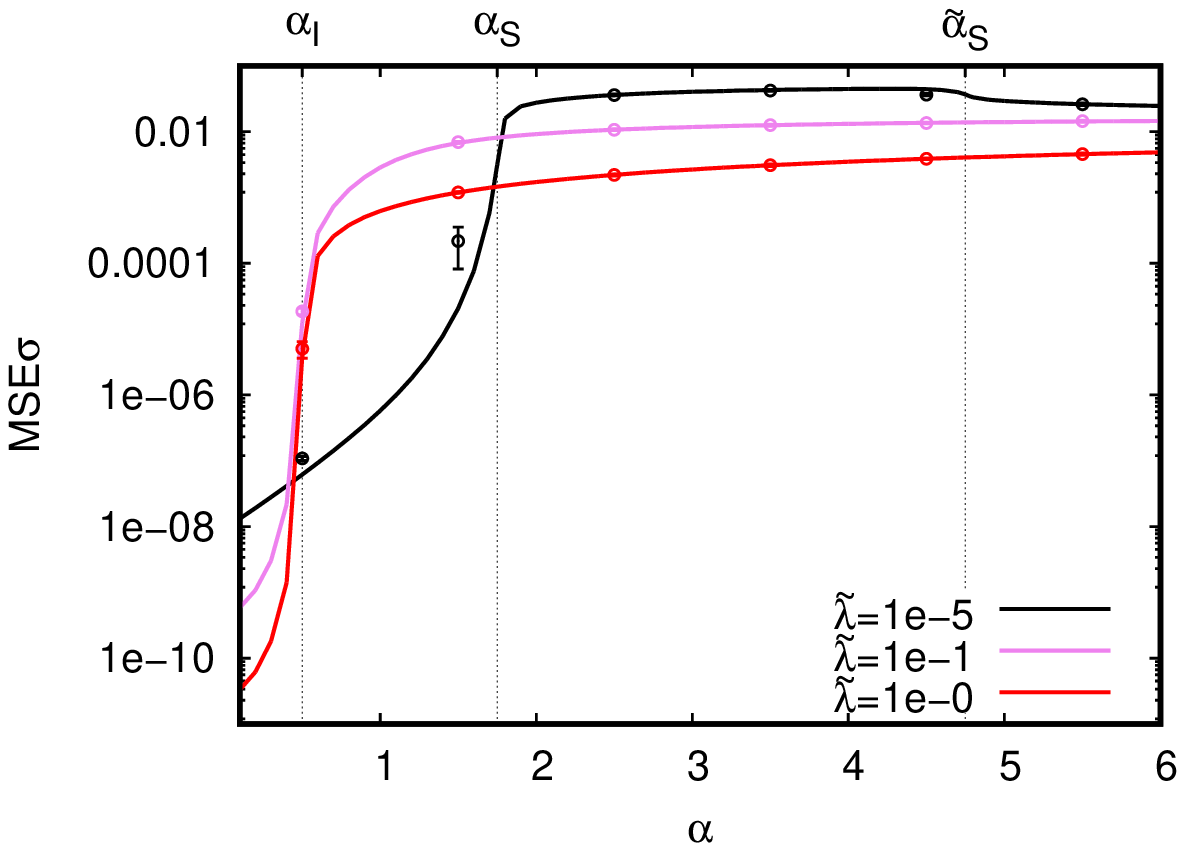}
\includegraphics[width=0.49\textwidth]{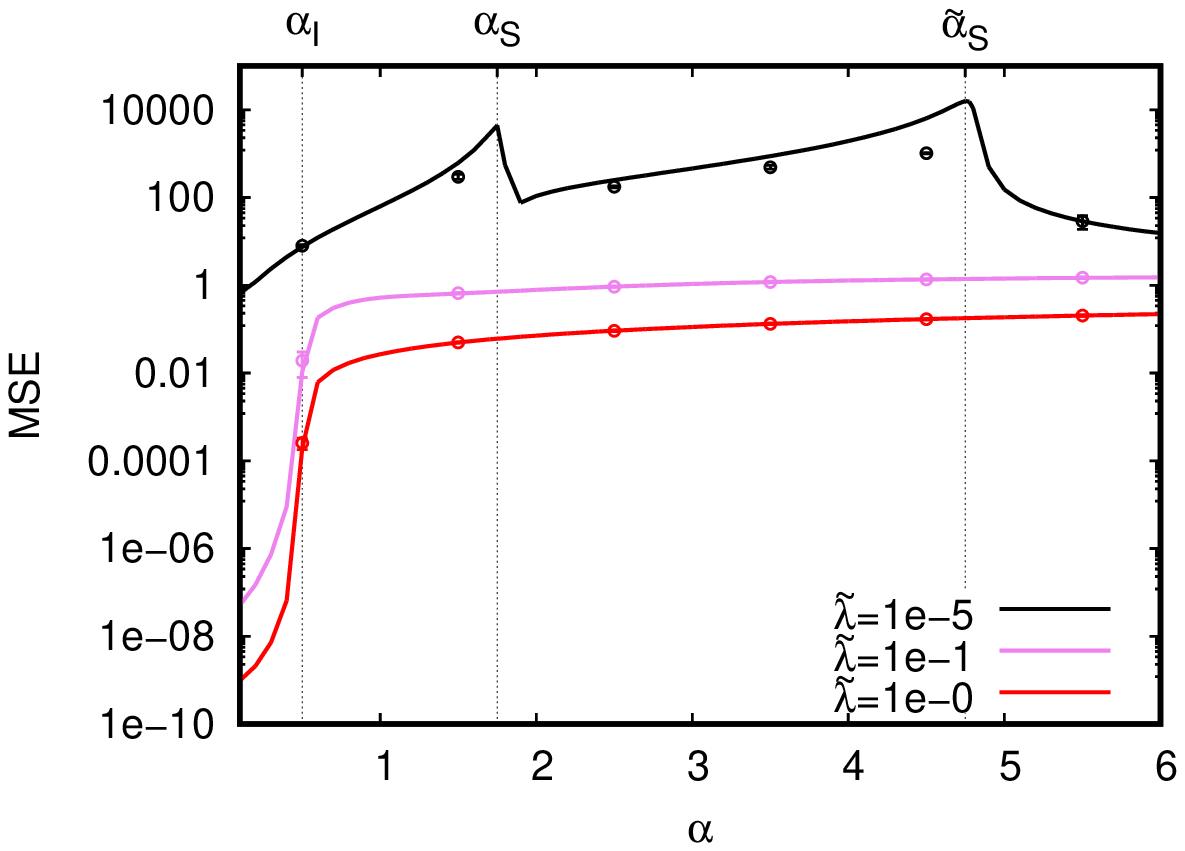}
\caption{ \label{fig:interpolation} \emph{Top left plot}: Typical student cross-entropy loss, averaged over the training set. \emph{Top right plot}: Typical student norm. \emph{Bottom left plot}: Typical MSE-distance between student and teacher outputs. \emph{Top right plot}: Typical MSE-distance between student and teacher preactivations. All the results are collected in the pure distillation setting ($\rho=0.2$, $\Delta=1$, $\chi=1$), at teacher regularizations $\tilde{\lambda}=1e-5$ (black), $\tilde{\lambda}=1e-1$ (violet), $\tilde{\lambda}=1$ (red). The data points with error bars represent the results of numerical experiments at $N=4000$ ($10$ samples per point). Due to the explosion of teacher and student norms (around the interpolation peak and the separability thresholds) in several occasions the employed numerical optimization routine (Adam optimizer \cite{kingma2014adam}) couldn't converge before the imposed hard cutoff of $2000$ epochs. This can explain the discrepancies with the theoretical predictions.}
\end{figure}

In order to understand the origin of the unusual interpolation peak at $\alpha=\eta$, in Fig.\,\ref{fig:interpolation}, we display the behavior of a series of relevant quantities: the average cross-entropy-per-pattern (top left), the student norm (top right), the average MSE-distance between teacher and student outputs (bottom left) and the average MSE between teacher and student preactivations. In all the plots we mark the three introduced thresholds, that in our parameter setting are located at $\alpha_I=0.5$, $\alpha_S\simeq1.75$ and $\tilde{\alpha}_S\simeq4.75$.

In the top left plot we can see that, when the student learns from a teacher with vanishing regularization (black curve), the optimal cross-entropy remains close to zero if the dataset is separable $\alpha<\alpha_S$, and jumps to finite values otherwise (similar to what usually happens in logistic regression: indeed, the weakly regularized teacher replicates the original labels almost exactly while $\alpha<\tilde{\alpha}_S$). The associated generalization error peak (cfr. the grey curve in Fig.\,\ref{fig:regularizations}) is thus caused by the explosion of the student norm (as expected with unregularized logistic regression at the separability threshold $\alpha=\alpha_S$).

On the other hand, at finite regularizations (violet and red curves) the teacher outputs are no longer binary and the minimum achievable cross-entropy becomes strictly greater zero, as can be seen in top right plot. While the number of associated linear constraints is lower than the number of parameters $\alpha<\alpha_I$, the student is able to faithfully reproduce the non-polarized teacher outputs (bottom left plot). However, by doing so the student overfits the noisy data and the increase in its norm (top right plot) gives rise to a sharp generalization error peak. These observations seem to be consistent with the scenario observed in \cite{phuong2019towards}, where it was shown that below the interpolation threshold the student converges to the projection of the teacher’s weight vector onto the data span and reproduces the teacher preactivations. Note that, when the teacher is over-regularized $\tilde{\lambda}=1$, teacher and student maximum norms are lower, partially tempering the generalization cusp. 

Finally, in the bottom right plot, we see the average MSE-distance between teacher and student preactivations. When the teacher is weakly regularized (black curve) the deviation between teacher and student preactivations sharply increases around the student interpolation threshold $\alpha_S$ (where the student norm is greater than the teacher's) and around the teacher interpolation threshold $\tilde{\alpha}_S$ (where the teacher norms reaches its maximum). At higher regularization levels this behavior disappears, since the cross-entropy minimization no longer induces great spikes in the student norm. 

\section{Numerical results for the balanced case}
\label{sec:balanced}

We here report the results obtained in the pure distillation setting when the two datapoints clusters are balanced $\rho=0.5$. As mentioned (and reported in \cite{mignacco2020role}), in this special case the ridge regularization is able to approach the Bayes optimal performance, in the $\lambda\to\infty$ limit.

\begin{figure}[ht]
\centering
\includegraphics[width=0.5\textwidth]{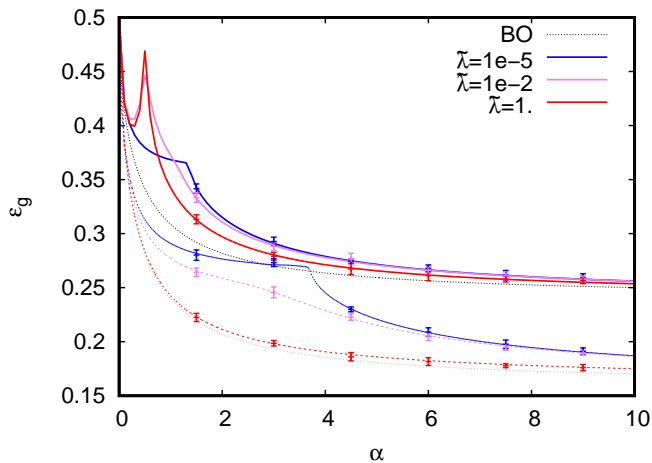}
\caption{\label{fig:balanced} Balanced clusters setting ($\rho=0.5$). (Black dashed line) Bayes optimal performance (e.g., achieved by the sparsified plug-in estimator). (Colored dashed lines) Teacher generalization performance, after training with ridge regularized logistic regression (colors indicate the regularizer intensity). (Full colored lines) Student generalization performance, in the pure distillation setting ($\lambda=1e-5$, $\chi=1$, $T=1$).}
\end{figure}
In Fig.\,\ref{fig:balanced}, we display the teacher (dashed lines) and student (full lines) generalization performances at varying ridge regularization intensity in the teacher loss, and compare them with the Bayes optimal performance (dashed). Clearly, higher regularization induces better generalization for the teacher and this property is directly inherited also by the student, when $\alpha$ is above the interpolation peak at $\alpha=\eta$ (more details in the main text).

\end{document}